\documentclass[sigconf]{acmart}
\AtBeginDocument{%
  }

\copyrightyear{2026}
\acmYear{2026}
\setcopyright{cc}
\setcctype{by}
\acmConference[WWW '26]{Proceedings of the ACM Web Conference 2026}{April 13--17, 2026}{Dubai, United Arab Emirates}
\acmBooktitle{Proceedings of the ACM Web Conference 2026 (WWW '26), April 13--17, 2026, Dubai, United Arab Emirates}
\acmPrice{}
\acmDOI{10.1145/3774904.3792376}
\acmISBN{979-8-4007-2307-0/2026/04}
\usepackage{lipsum}
\usepackage{graphicx}
\usepackage{hyperref}
\usepackage{algorithm}
\usepackage{algorithmic}
\usepackage{nicematrix,booktabs}
\usepackage{multirow}
\usepackage{ragged2e}
\usepackage{tabularx}
\usepackage{tcolorbox}
\usepackage{caption}
\usepackage{subfigure}
\usepackage{transparent}
\usepackage{listings}
\usepackage{multicol}
\usepackage{boldline}
\usepackage{arydshln}
\usepackage{diagbox}
\usepackage{amssymb}
\usepackage{pifont}
\newcommand{\cmark}{\ding{51}}
\newcommand{\xmark}{\ding{55}}
\usepackage{tikz}
\usepackage{enumitem}
\usepackage{changepage}
\usepackage{multirow}
\usepackage{soul}
\usepackage{xcolor}
\setlist[itemize]{leftmargin=*}

\tcbset{
  aibox/.style={
    width=\columnwidth,
    top=10pt,
    colback=gray!10,
    colframe=gray,
    colbacktitle=gray,
    center,
  }
}
\newtcolorbox{AIbox}[2][]{aibox,title=#2,#1}

\definecolor{gblue}{RGB}{66,133,244}
\definecolor{gred}{RGB}{219,68,55}
\definecolor{gyellow}{RGB}{244,180,0}
\definecolor{ggreen}{RGB}{15,157,88}
\definecolor{lpcolor}{RGB}{42,74,138}
\definecolor{morelcolor}{RGB}{185,18,32}
\definecolor{bgcolor}{RGB}{230,245,208}
\definecolor{framecolor}{RGB}{244,109,67}
\definecolor{mulberry}{rgb}{0.77, 0.29, 0.55}

\definecolor{lightblue}{RGB}{212, 235, 255}
\definecolor{grey1}{RGB}{96, 101, 102}
\definecolor{lightorange}{RGB}{255, 204, 168}
\definecolor{lightyellow}{RGB}{255, 255, 168}
\definecolor{lightgreen}{RGB}{224, 242, 213}
\definecolor{lightred}{RGB}{249,202,202}
\definecolor{lightgray}{RGB}{230,230,230}
\definecolor{deepred}{RGB}{152, 1, 0}

\newtcbox{\hlprimarytab}{on line, rounded corners, box align=base, colback=lightgreen,colframe=white,size=fbox,arc=3pt, before upper=\strut, top=-2pt, bottom=-4pt, left=-2pt, right=-2pt, boxrule=0pt}

\newtcbox{\hlsecondarytab}{on line, box align=base, colback=lightred,colframe=white,size=fbox,arc=3pt, before upper=\strut, top=-2pt, bottom=-4pt, left=-2pt, right=-2pt, boxrule=0pt}

\newtcbox{\hlwarningtab}{on line, box align=base, colback=lightyellow, colframe=white, size=fbox, arc=3pt, before upper=\strut, top=-2pt, bottom=-4pt, left=-2pt, right=-2pt, boxrule=0pt}

\newcommand{\dshifted}{\raisebox{0.5\depth}{$\downarrow$}}
\newcommand{\ushifted}{\raisebox{0.5\depth}{$\uparrow$}}

\newcommand{\dg}[1]{{\hlprimarytab{\dshifted{#1}}}}

\newcommand{\dr}[1]{{\hlsecondarytab{\dshifted{#1}}}}
\newcommand{\ug}[1]{{\hlprimarytab{\ushifted{#1}}}}

\definecolor{c1}{cmyk}{0,0.6175,0.8848,0.1490} 
\definecolor{c2}{cmyk}{0.1127,0.6690,0,0.4431}
\definecolor{c3}{cmyk}{0.3081,0,0.7209,0.3255} 
\definecolor{c4}{cmyk}{0.6765,0.2017,0,0.0667}
\definecolor{c5}{cmyk}{0,0.8765,0.7099,0.3647} 
\definecolor{forestgreen}{HTML}{397727}

\definecolor{backcolour}{rgb}{0.95,0.95,0.92}
\definecolor{codegreen}{rgb}{0,0.6,0}
\definecolor{mygreen}{HTML}{88EABB}
\definecolor{OliveGreen}{HTML}{00693E}
\definecolor{markgreen}{rgb}{0.3, 0.73, 0.09}
\definecolor{markred}{rgb}{0.8, 0.0, 0.0}
\definecolor{deepred}{RGB}{152, 1, 0}

\newcommand{\ds}{\textit{VisualTimeAnomaly}}
\newcommand{\m}{\textit{TSAD-Agents}}

\begin{document}

\title{Can Multimodal LLMs Perform Time Series Anomaly Detection?}

\author{Xiongxiao Xu}
\affiliation{%
  \institution{Illinois Institute of Technology}
  \city{Chicago}
  \state{IL}
  \country{USA}
}
\email{xxu85@hawk.illinoistech.edu}

\author{Haoran Wang}
\affiliation{%
  \institution{Emory University}
  \city{Atlanta}
  \state{GA}
  \country{USA}
}
\email{haoran.wang@emory.edu}

\author{Yueqing Liang}
\affiliation{%
  \institution{Illinois Institute of Technology}
  \city{Chicago}
  \state{IL}
  \country{USA}
}
\email{yliang40@hawk.illinoistech.edu}

\author{Philip S. Yu}
\affiliation{%
  \institution{University of Illinois Chicago}
  \city{Chicago}
  \state{IL}
  \country{USA}
}
\email{psyu@uic.edu}

\author{Yue Zhao}
\affiliation{%
  \institution{University of Southern California}
  \city{Los Angeles}
  \state{CA}
  \country{USA}
}
\email{yue.z@usc.edu}

\author{Kai Shu}
\affiliation{%
  \institution{Emory University}
  \city{Atlanta}
  \state{GA}
  \country{USA}
}
\email{kai.shu@emory.edu}

\renewcommand{\shortauthors}{Xiongxiao Xu et al.}

\begin{abstract}
Time series anomaly detection (TSAD) has been a long-standing pillar problem in Web-scale systems and online infrastructures, 
such as service reliability monitoring, system fault diagnosis, and performance optimization.
Large language models (LLMs) have demonstrated unprecedented capabilities in time series analysis, the potential of multimodal LLMs (MLLMs), particularly vision-language models, in TSAD remains largely under-explored. 
One natural way for humans to detect time series anomalies is through visualization and textual description. It motivates our research question: \textit{Can multimodal LLMs perform time series anomaly detection?} 
Existing studies often oversimplify the problem by treating point-wise anomalies as special cases of range-wise ones or by aggregating point anomalies to approximate range-wise scenarios.
They limit our understanding for realistic scenarios such as multi-granular anomalies and irregular time series. To address the gap, we build a {\ds} benchmark to comprehensively investigate zero-shot capabilities of MLLMs for TSAD, progressively from point-, range-, to variate-wise anomalies, and extends to irregular sampling conditions.
Our study reveals several key insights. 1) MLLMs and traditional TSAD methods are complementary: MLLMs excel at coarse-grained anomalies while traditional methods are effective at fine-grained anomalies.
2) MLLMs are resilient to irregular time series.
3) Input time series modality changing from text to image makes information focus shift from quantitative variations to qualitative patterns while significantly reducing hallucinations.
Built on the findings, we propose a MLLMs-based multi-agent framework {\m} to achieve automatic TSAD. Our framework comprises scanning, planning, detection, and checking agents that synergistically collaborate to reason, plan, and self-reflect to enable automatic TSAD. These agents adaptively invoke tools such as traditional methods and MLLMs and dynamically switch between text and image modalities to optimize detection performance. 
\end{abstract}





\maketitle

\begin{figure}[t!]
    \centering
    \includegraphics[width=0.5\textwidth]{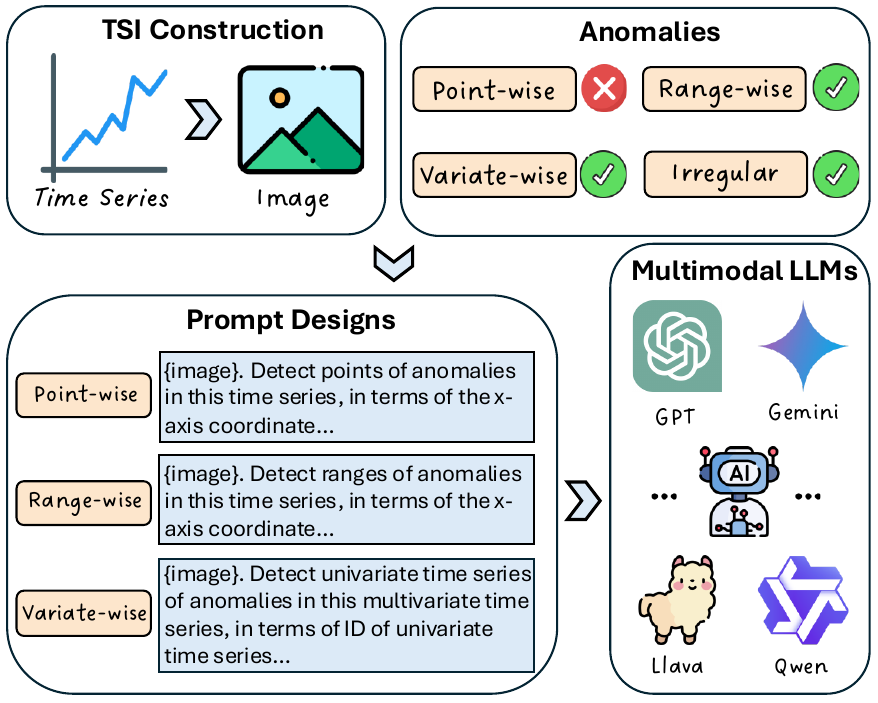}
     \caption{The workflow of {\ds}. Time series numerical data is converted into the image format. Then the textual description and time series images (TSIs) are fed into MLLMs for anomaly detection.
     }
     \vspace{-0.5cm}
    \label{fig:workflow}
\end{figure}

\section{Introduction}
As Web and IoT systems grow increasingly complex and interconnected, detecting anomalies in time series data plays a pivotal role in ensuring the stability and efficiency of large-scale online infrastructures. These systems continuously produce extensive streams of time series signals, such as CPU and memory utilization, latency traces and throughput logs, and user traffic and sensor telemetry. Timely detection of anomalous resource usage or traffic surges can prevent cascading failures and sustain a seamless user experience. Consequently, time series anomaly detection (TSAD) serves as a cornerstone of AIOps pipelines and Web observability frameworks.

LLMs have gained popularity in diverse applications, including TSAD~\cite{liu2024large}, due to their attractive zero-shot in-context learning capability~\cite{yang2024ad}.
The existing studies in language models for TSAD directly regard time series numerical data as text~\cite{dong2024can}. However, they may be not be optimal~\cite{alnegheimish2024large} as humans detecting time series anomalies is typically by a combination of visualization and languages. For example, AIOps practitioners~\cite{li2022constructing} monitor visual dynamics of metrics like CPU usage over time and correlate them with textual information to check system conditions. This practice aligns with the development of multimodal LLMs (MLLMs), which evolve from processing text-only to hybrid text-image modalities. Accordingly, we pose a research question: \textit{Can multimodal LLMs perform time series anomaly detection?}

Previous studies have shown benefits of visualizing time series for anomaly detection~\cite{zhou2025can,zhuang2024see}. However, they either treat point-wise anomalies as a special case of range-wise anomalies or reduce range-wise anomalies to a trivial aggregation of point-wise ones (Table~\ref{tab:related}). They restrict our knowledge away from more complex and realistic scenarios, including variate-wise anomalies and irregular time series. In real-world environments, time series data are often collected under unstable conditions with unexpected interruptions, resulting in imperfect and irregular observations. 
For example, in large-scale cloud services, system metrics such as CPU utilization or network latency can be intermittently missing due to node failures or delayed telemetry updates, while in edge or mobile IoT systems, sensor readings may arrive at uneven intervals because of limited connectivity or energy constraints.
In such cases, point-, range-, and variate-wise anomalies may all occur alongside temporal irregularities. Moreover, while previous work reports visual advantages of time series than text, there remains limited analysis on the underlying factors that drive performance differences across modalities. To bridge these gaps, this paper investigates zero-shot capability of MLLMs in TSAD by addressing the following questions:
\begin{itemize}
    \item \textbf{RQ1:} Can MLLMs perform zero-shot multi-granular (point, range, and variate) time series anomaly detection?
    \item \textbf{RQ2:} Can MLLMs detect irregular time series anomalies in a zero-shot setting? 
    \item \textbf{RQ3:} Which input modality (text, image or hybrid) best supports time series anomaly detection with MLLMs? 
\end{itemize}
To address the above questions, we introduce the {\ds} benchmark to systematically evaluate zero-shot capability of MLLMs in TSAD, as illustrated in Figure~\ref{fig:workflow}. Specifically, we investigate 2 types of point-, 3 types of range-, and 4 types of variate-wise time series anomalies~\cite{lai2021revisiting,zamanzadeh2024deep} (RQ1). To explore robustness to irregularity, we further introduce random points dropping at varying rates (RQ2). MLLMs can process time series in three primary modalities~\cite{xu2025beyond}: text (via prompt-based encoding), images (via visual transformation), or both. To explore which modality best supports anomaly detection, we design a series of controlled experiments (RQ3). 
Overall, {\ds} serves as a configurable benchmark for advancing multimodal TSAD research. It supports flexible generation of diverse multimodal anomaly detection datasets by varying key parameters such as anomaly granularity, irregularity ratio, series length, and dimensionality. In our study, we apply {\ds} to benchmark representative MLLMs. 
\\\textbf{Key Takeaways.} Our work uncovers several key insights:
\begin{itemize}
    \item \textbf{MLLMs and traditional methods are complementary regarding to anomaly granularities}. MLLMs are superior in coarse-grained (range- and variate-wise) anomalies while traditional methods excel at fine-grained (point-wise) anomalies.
    \item \textbf{MLLMs are resilient to irregularity by visualizing time series as images}, even with 25\% of the data missing, where traditional TSAD methods struggle to handle.
    \item Changing the input modality \textbf{from text to image shifts MLLMs' focus from quantitative variations to qualitative patterns}. This change also significantly reduces MLLMs' hallucinations due to the reduced input length. 
\end{itemize}

Built on these findings, we propose a multi-agent framework {\m}, which not only leverage the complementary strengths of traditional methods and MLLMs with different modalities, but also comprehensively harnesses MLLMs' internal capabilities such as reasoning, planning, tool-use, and self-reflection, to enable automatic TSAD with minimal human intervention. The framework comprises four collaborative agents: Scanning, Detection, Planning, and Checking agents. They synergistically coordinate to automatically reason the anomaly type, choose appropriate tools for precise anomaly detection, and refine their predictions, while adaptively switching between textual and visual input modalities. To the best of our knowledge, {\ds} is a early benchmark and {\m} is the first multi-agent system designed for multimodal TSAD. The code is here\footnote{https://github.com/mllm-ts/VisualTimeAnomaly}. We summarize our contributions:
\begin{itemize}
    \item We build an \textbf{early benchmark {\ds} for multimodal TSAD}, covering point-, range-, variate-wise, and irregular anomalies. The benchmark is highly configurable to facilitate future research in this area.
    \item We discover several \textbf{actionable insights} that advance the understanding of MLLMs in TSAD.
    \item We propose \textbf{the first multi-agent systems {\m} for multimodal TSAD}. Guided by the insights, {\m} harnesses MLLMs’ strong reasoning, planning, tool-use, and self-reflection capabilities to achieve automatic TSAD.
\end{itemize}

\begin{table}[!t]
\renewcommand{\arraystretch}{1.2}%
    \centering
    \setlength{\tabcolsep}{4pt}
    \resizebox{1.00\linewidth}{!}{%
        \begin{Tabular}{lccccc}
        \hline
        
        \textbf{Work} & \textbf{Point} & \textbf{Range} & \textbf{Variate} & \textbf{Irr.} & \textbf{Multi-Agents}\\ \hline
        
        \cite{zhou2025can} & \textcolor{markgreen}{\cmark} & \textcolor{markgreen}{\textbf{\cmark}} & \textcolor{markred}{\xmark} & \textcolor{markred}{\xmark} & \textcolor{markred}{\xmark} \\

        \cite{zhuang2024see} & \textcolor{markgreen}{\cmark} & \textcolor{markgreen}{\cmark} & \textcolor{markred}{\xmark} & \textcolor{markred}{\xmark} & \textcolor{markred}{\xmark} \\
        
        \textbf{Ours} & \textcolor{markgreen}{\textbf{\cmark}} & \textcolor{markgreen}{\textbf{\cmark}} & \textcolor{markgreen}{\textbf{\cmark}} & \textcolor{markgreen}{\cmark} & \textcolor{markgreen}{\cmark} \\ \hline
        \end{Tabular}
    }
\caption{Comparison between our work and existing studies leveraging MLLMs for TSAD. Our work considers multi-granular and irregular time series anomalies, and also build a multi-agent framework to automate efficient TSAD.
}
\label{tab:related}
\vspace{-0.7cm}
\end{table}

\begin{figure*}[t!]
    \centering
    \subfigure[Point-wise and range-wise anomalies]{
        \begin{minipage}{0.49\textwidth}
            \includegraphics[width=1\textwidth]{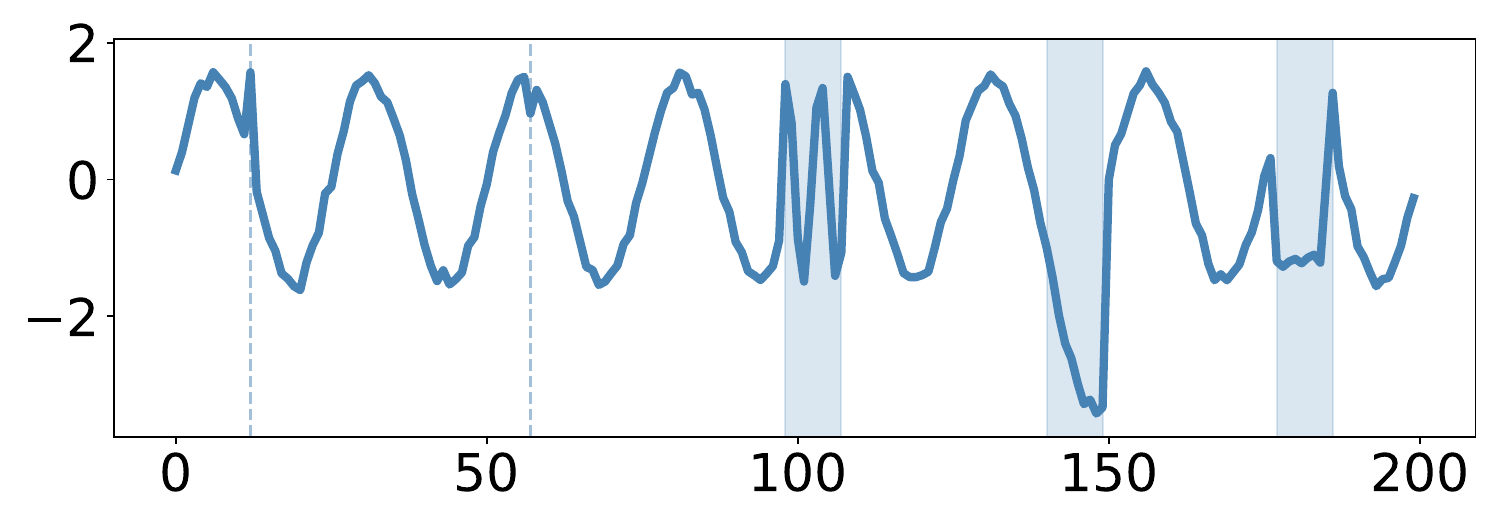}
        \end{minipage}\label{fig:univariate_anomaly}
    }
    \hspace{-0.3cm}
    \subfigure[Variate-wise anomalies]{
        \begin{minipage}{0.49\textwidth}
            \includegraphics[width=1\textwidth]{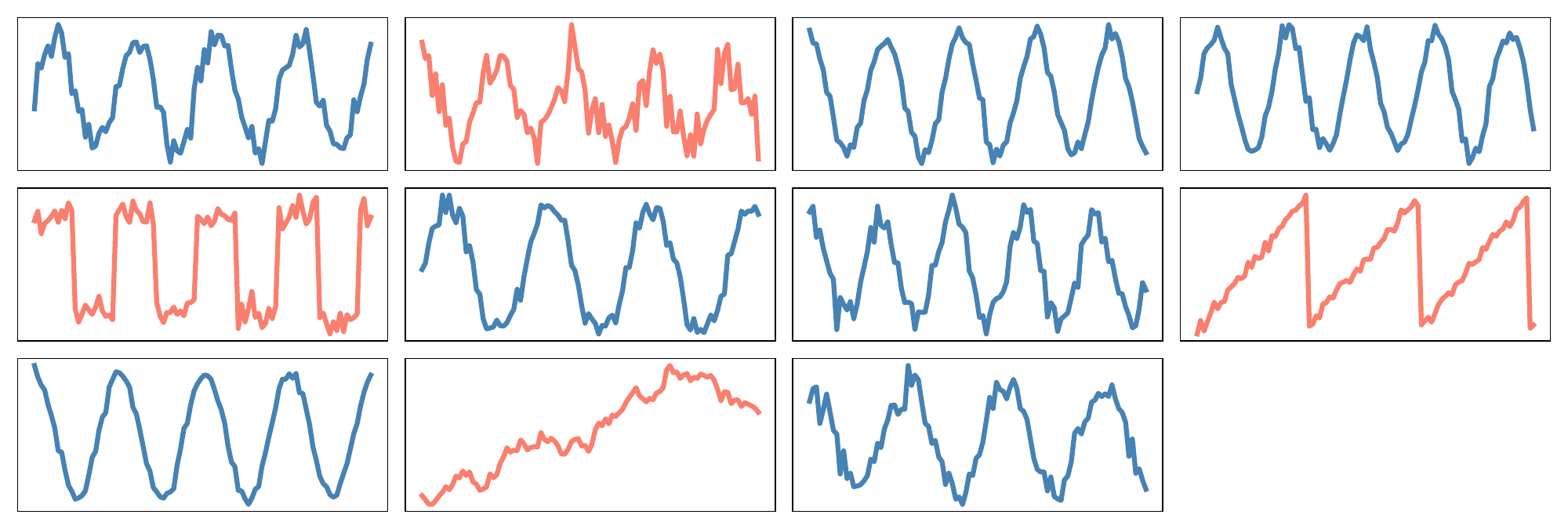}
        \end{minipage}\label{fig:multivariate_anomaly}
    }
    \caption{The illustration of point-, range-, and variate-wise anomalies. \textbf{Left}: From left to right, dashed lines and highlighted intervals represent global, contextual, seasonal, trend, and shapelet anomalies. \textbf{Right}: From left to right and top to bottom, time series marked by the red color indicate triangle, square, sawtooth, and random anomalies.}
    \label{fig:anomaly}
\end{figure*}

\section{{\ds} Benchmark}\label{sec:anomaly_definition}
In this section, we present the {\ds} benchmark, including anomaly definition, datasets, and other details.

\subsection{Time Series Anomaly Definition}
We formally define time series anomalies in different granularities and introduce their definitions in irregular anomalies.
\\\textbf{Point- and range-wise anomalies} are within univariate time series. Formally, \textit{univariate time series} can be denoted as $\mathbf{x}=\{x_1,x_2,\dots,x_T\}$ with $T$ time points, where $x_t\in\mathbb{R}$ represents value at time $t$. 

\textit{Point-wise anomalies} are unexpected incidents at individual time points~\cite{lai2021revisiting}:
\begin{equation}\label{eq:point_anomaly}
    |x_t-\hat{x_t}| > \delta
\end{equation}
where $\hat{x_t}$ is the expected value at time $t$, which can be the global mean value or mean value of a context window, and $\delta$ is the threshold. 
Accordingly, we have 2 types of point-wise anomalies: global anomalies and contextual anomalies (Figure~\ref{fig:univariate_anomaly}). 

\textit{Range-wise anomalies} are anomalous subsequences characterized by changes in seasonality, trend, or shapelet~\cite{lai2021revisiting}. Formally, within a time series $\mathbf{x}$, a subsequence $\mathbf{x}_{i:j}$ from time $i$ to $j$ can be considered anomalous if:
\begin{equation}\label{eq:context_anomaly}
    diss(\mathbf{x}_{i,j}, \hat{\mathbf{x}}_{i,j}) > \delta
\end{equation}
where $diss$ is a function measuring the dissimilarity between two subsequences, such as dynamic time warping~\cite{berndt1994using}, and $\hat{\mathbf{x}}_{i,j}$ is the expected subsequence regarding to seasonality, trend or shapelet. Accordingly, we define 3 types of range-wise anomalies: seasonal, trend, and shapelet anomalies (Figure~\ref{fig:univariate_anomaly}).
\begin{figure*}[t!]
    \centering
    \includegraphics[width=\textwidth]{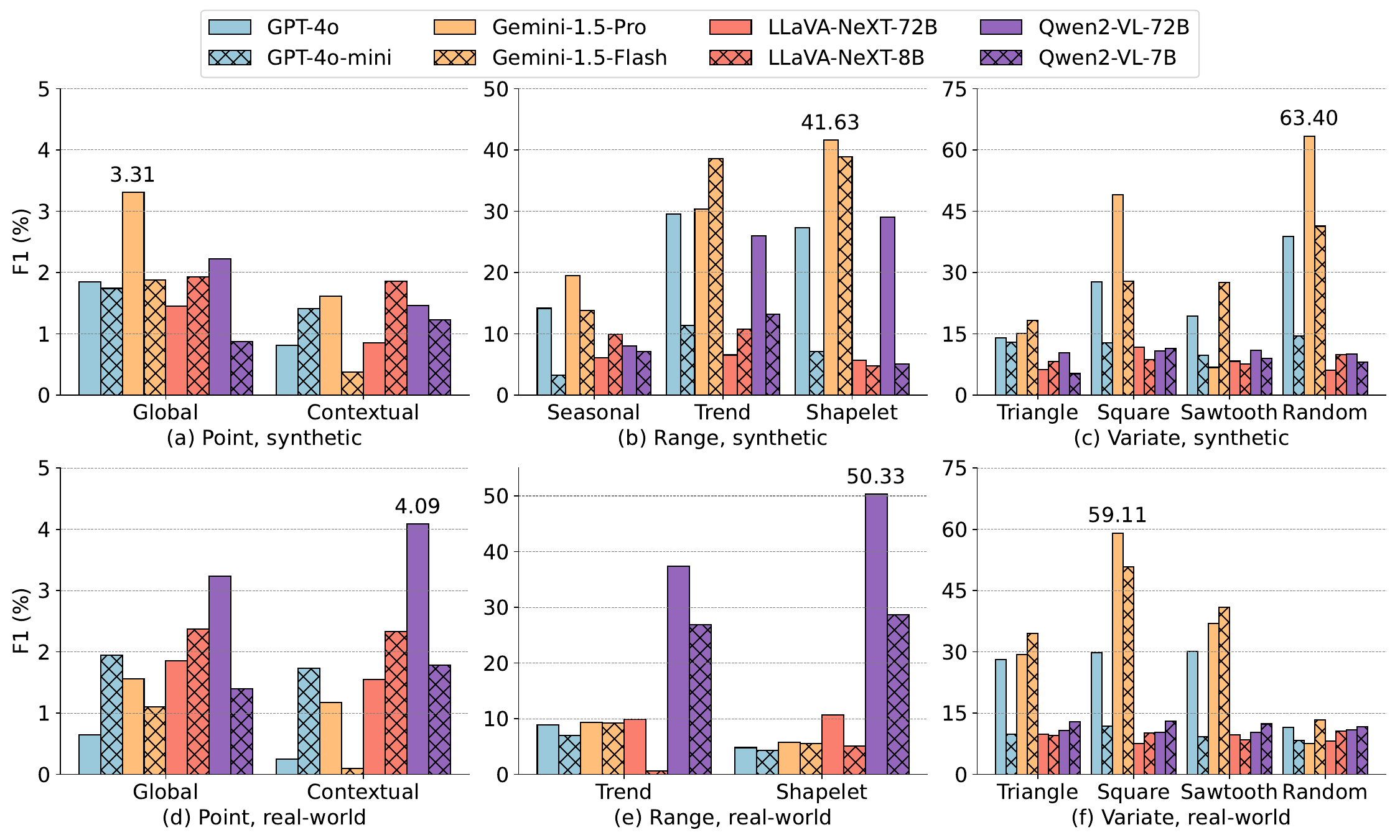}
     \caption{The detection performance of MLLMs across point-, range-, and variate-wise anomalies on both synthetic (top) and real-world (bottom) time series. 
     From left to right, F1 scores progressively improve as anomaly granularity changes from fine-grained (point-wise) to coarse-grained (range- and variate-wise).
     }
     \vspace{-0.3cm}
    \label{fig:bar_point_range_variate}
\end{figure*}
\\\textbf{Variate-wise anomalies} are defined within multivariate time series which simultaneously captures multiple temporal variables. Formally, a \textit{multivariate time series} can be denoted as $\mathbf{X}=\{\mathbf{x}^1,\mathbf{x}^2,...,\mathbf{x}^M\}$ with $M$ variables, where $\mathbf{x}^m\in\mathbb{R}^T$ is a $T$-dimensional univariate vector for the $m^{th}$ variate.

\begin{figure}[b!]
    \vspace{-0.5cm}
    \centering
    \includegraphics[width=0.48\textwidth]{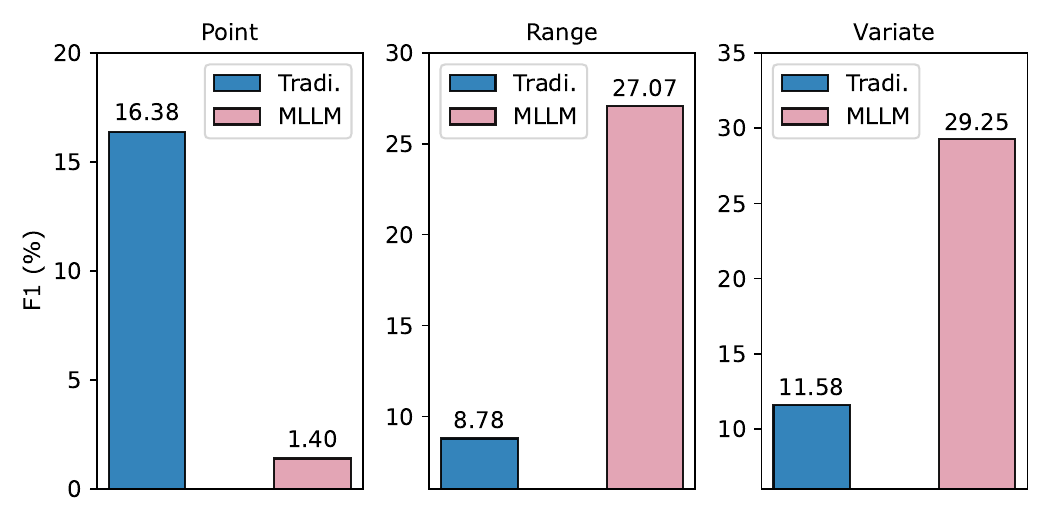}
     \caption{Performance comparison of MLLMs and traditional methods (\textit{Tradi.}). 
     MLLMs excel at range- and variate-wise anomalies while struggling at point-wise anomalies compared to traditional TSAD methods.
     }
     \vspace{-0.5cm}
    \label{fig:bar_comparison_tsad}
\end{figure}

\begin{figure*}[t!]
    \centering
    \includegraphics[width=\textwidth]{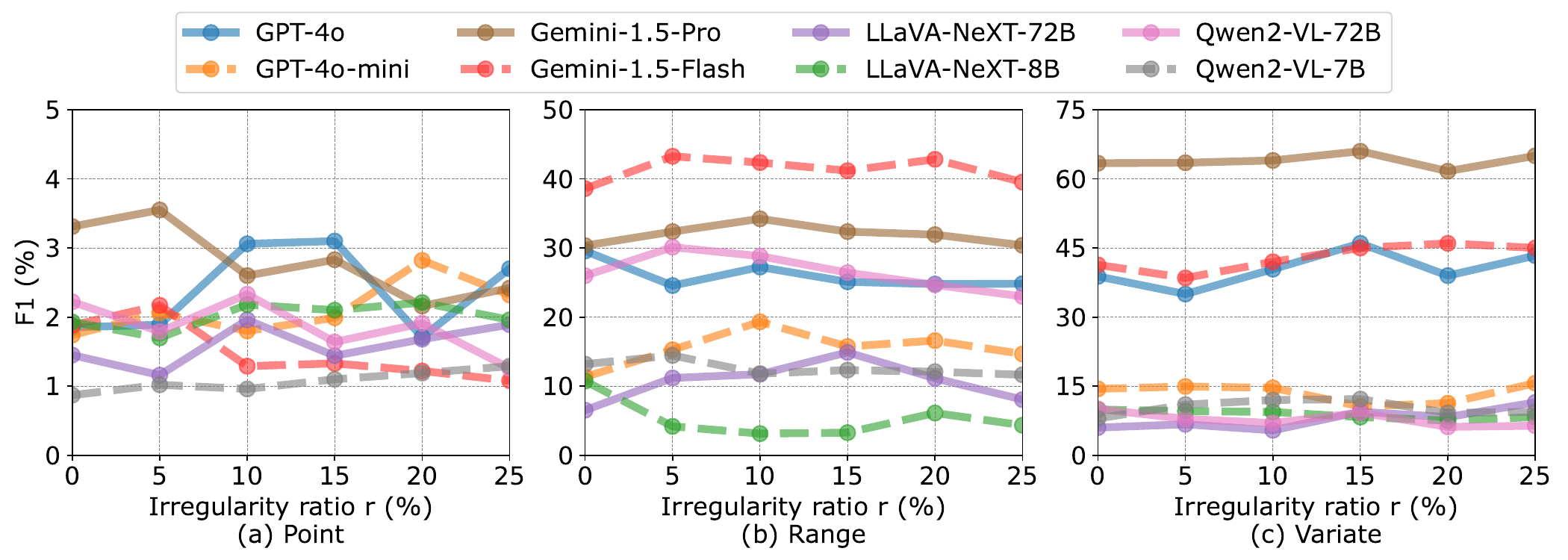}
     \caption{Impact of the irregularity ratio $r$ on MLLMs for point-wise, range-wise, and variate-wise anomalies. It indicates that MLLMs are robust to the irregularity in time series.}
    \label{fig:irregular}
\end{figure*}

\textit{Variate-wise anomalies} are entire time series from certain variates which significantly deviate from others within a multivariate time series~\cite{zamanzadeh2024deep}. We assume each variate $m$ is governed by a generative function $G(x)_m$. We define variate-wise anomalies as follows:
\begin{equation}\label{eq:context_anomaly}
    diss(G(x)_m, \hat{G}(x)_m) > \delta
\end{equation}
where the $diss$ function measures dissimilarity between two univariate time series, and $\hat{G}(x)_m$ denotes the expected generative function, or the majority generative functions among all variates, such as \textit{sine} or \textit{cosine} waves. Figure~\ref{fig:multivariate_anomaly} illustrates 4 types of variate-wise anomalies~\cite{kraft2017lp}: triangle, square, sawtooth, and random anomalies.
\\\textbf{Irregular anomalies} are within irregular time series which refer to univariate/multivariate sequences with with non-uniform or missing sampling intervals. 
Formally, an \textit{irregular univariate time series} can be represented as $\mathbf{x}_{irr}=\{x_1,...,x_{i-1},x_{i+2}...,x_T\}$ with $S$ $(S<T)$ observed points, where the $i$-th observation is missing. 
Similarly, an \textit{irregular multivariate time series} is denoted as $\mathbf{X}_{irr}=\{\mathbf{x}_{irr}^1,\mathbf{x}_{irr}^2,...,\mathbf{x}_{irr}^M\}$, where each $\mathbf{x}_{irr}^m$ has $S$ observed points for the $m^{th}$ variate. We define the irregularity ratio for irregular univariate/multivariate time series as $r=1-\frac{S}{T}$.

\textit{Irregular anomalies} extend the definitions of anomalies from regular to irregular time series. They are generated by randomly removing data points from regular sequences while preserving the underlying anomaly patterns. Contextual anomalies are excluded from this construction because removing data points in their vicinity would distort the temporal context required for their definition.

\vspace{-0.4cm}
\subsection{Benchmark Setting}
Our benchmark scripts are extended from~\cite{lai2021revisiting} which serves as a foundation work for point- and range-wise anomalies. 
\\\textbf{Datasets.}
We inject anomalies into synthetic time series (\textit{sine}/\textit{cosine} waves)~\cite{zhou2025can}. The length of univariate time series and multivariate sequences are set as 400 (same as \cite{lai2021revisiting}) and 200 (reduced length due to added dimensionality). Additionally, we introduce anomalies into two real-world datasets from the UCR and UEA archive~\cite{dau2019ucr,bagnall2018uea} as a real-world case study. The Symbol dataset is as univariate series and the ArticularyWordRecognition dataset is as multivariate because their lengths (398 and 144) are close to the synthetic setting. 
Each anomaly type is generated with 100 TSIs with different noises. All experiments are run on the 100 images 3 times. 
We do not use off-the-shelf time series anomaly detection dataset~\cite{zhou2025can} due to: (1) need for specific anomaly type isolation, and (2) known label quality issues in existing public dataset~\cite{wu2021current}.
\\\textbf{Baselines.}
The baselines include both classical TSAD methods, iForest~\cite{liu2008isolation} and OCSVM~\cite{scholkopf1999support}, and deep learning methods, OmniAnomaly~\cite{su2019robust}, THOC~\cite{shen2020timeseries}, and TranAD~\cite{tuli2022tranad}
\\\textbf{Multimodal LLMs.} We employ representative multimodal LLMs, including both proprietary (GPT-4o and Gemini-1.5) and open-source (LLaVA-NeXT and Qwen2-VL) models. 
For each MLLM type, we employ one larger and one lightweight variant to ensure a comprehensive assessment.
\\\textbf{Evaluation Metrics.} We employ precision, recall, and F1 as evaluation metrics following community~\cite{lai2021revisiting,zamanzadeh2024deep,zhou2025can}.

\vspace{-0.2cm}
\section{RQ1: Multi-Granular TSAD}
To answer RQ1, we benchmark MLLMs across different anomaly granularities. One challenge is to fairly compare variate- anomalies with point- and range-wise, as they exist in multivariate and univariate series, respectively. Following prior work~\cite{lai2021revisiting}, we inject anomalous points or ranges of length 20 into univariate time series of length 400. To maintain a comparable $5\%$ anomaly ratio, one variate-wise anomaly is injected into a 20-variate time series.

\textbf{Multi-Granular Detection Results.}
Figure~\ref{fig:bar_point_range_variate} shows that \textbf{MLLMs are more effective at detecting coarse-grained anomalies than fine-grained anomalies}. Specifically, for point-wise anomalies, the maximum F1 scores across MLLMs is 4.09 while they are 50.33 for range-wise anomalies and 63.40 for variate-wise anomalies. The ranking of detection effectiveness of MLLMs is: \textit{Variate}$>$\textit{Range}$>$\textit{Point}. It indicates that MLLMs respond more sensitive to coarse-grained patterns than fine-grained variations. 

\textbf{Comparison with Traditional TSAD Methods.}
We further investigate the role of MLLMs in TSAD by comparing them with traditional TSAD baselines. Figure~\ref{fig:bar_comparison_tsad} shows the average performance comparison of MLLMs and baselines, revealing that \textbf{MLLMs and traditional TSAD methods are complementary regarding to anomaly granularities}. In detail, MLLMs are more effective at coarse-grained anomalies (i.e., range and variate), whereas traditional TSAD methods are better at fine-grained (i.e., point) anomalies. This is due to the limited numerical reasoning ability of MLLMs, such as incorrectly evaluating 9.11 $>$ 9.9~\cite{yang2024number}, causing them insensitive to point-wise changes.

\vspace{-0.2cm}
\section{RQ2: Irregular TSAD}
To explore RQ2, we vary the irregularity ratio $r$ from $0\%$ to $25\%$ in increments of $5\%$, focusing on three anomaly types: global (point-), trend (range-), and random (variate-wise). These types are selected based on either their strong performance or low variance.

\textbf{Irregular Detection Results.} Figure~\ref{fig:irregular} presents the impact of the irregularity ratio $r$ on MLLMs' detection performance. We clearly observe that \textbf{MLLMs are robust against irregularity in time series.} For example, on range-wise anomalies, F1 score of Gemini changes from 30.34 to 30.39 with $r=0\%$ to $r=25\%$. This result underscores the high resilience of MLLMs to handle irregular time series. Moreover, we notice that the fluctuation of MLLMs on point-wise anomalies is slightly higher than them on coarser-grained anomalies. It is consistent with our previous insight that MLLMs are inferior in fine-grained anomalies.

\textbf{Comparison with Traditional TSAD Methods.}
We compare the performance of MLLMs against traditional TSAD approaches on irregular time series to assess the position of MLLMs within TSAD. However, classical TSAD baselines typically assume regularly sampled data and often fail or degrade substantially when encountering temporal irregularities. For instance, OCSVM and TranAD can raise runtime errors when handling inputs with uneven timestamps. To ensure compatibility, we adopt a common preprocessing strategy \textit{mean imputation}, which fills in missing entries with the average of the corresponding univariate series.

\begin{figure*}[t]
    \centering
    \includegraphics[width=\textwidth]{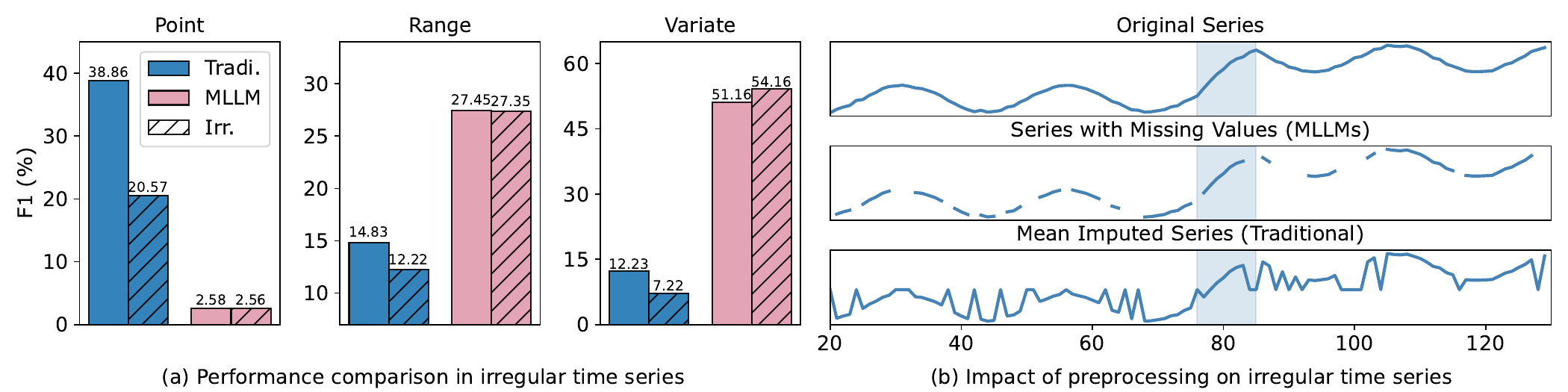}
     \caption{\textbf{Left}: Performance comparison of MLLMs and traditional methods (\textit{Tradi.}) on irregular time series (\textit{Irr.}). 
     MLLMs remain robust compared to traditional methods. \textbf{Right}: Visualization for a trend anomaly in irregular time series. The highlighted region marks the anomaly interval. MLLMs operate directly on incomplete data, preserving the overall structure.
     }
     \vspace{-0.3cm}
    \label{fig:bar_comparsion_irr}
\end{figure*}

Figure~\ref{fig:bar_comparsion_irr}(a) reports the average F1 scores of MLLMs and traditional methods when anomaly detection conditions shift from regular ($r=0\%$) to irregular ($r=25\%$) time series. The results indicate that \textbf{MLLMs demonstrate strong resilience to irregularity, while traditional methods struggle to maintain performance.} Specifically, for point-wise anomalies, traditional methods suffer a substantial F1 drop (from 38.86 to 20.57 with a 47\% decrease), whereas MLLMs maintain stable performance (from 2.58 to 2.56 with only a 0.78\% decrease). To gain deeper understanding into the robustness, Figure~\ref{fig:bar_comparsion_irr}(b) visualizes the impact of preprocessing.  We illustrate three versions of the same time series: the original series, the irregular version processed as an image (used by MLLMs), and the mean-imputed version (used by traditional methods). The visualization-based approach adopted by MLLMs naturally encodes missing values as blank regions, avoiding the distortions introduced by imputation. This enables MLLMs to preserve critical structural features, such as trends and shapelets, and localize anomalies without requiring explicit temporal alignment.

\begin{table}[t]
    \centering
    \scalebox{0.85}{
    \begin{tabular}{lcccc}
        \toprule
        & \textbf{Condition 1} & \textbf{Condition 2} & \textbf{Condition 3} \\
        \cmidrule(r){2-2} \cmidrule(lr){3-3} \cmidrule(l){4-4}
        \textbf{Method} & ViTST & MLLM (Text) & MLLM (Vision) \\
        \textbf{Metrics} & Accuracy (\%) & F1 (\%) & F1 (\%) \\
        \midrule
        Point & 43.7$\approx$50.0 & 5.88$\rightarrow$4.77\dr{18.8\%} & 2.58$\rightarrow$2.56\dr{0.77\%} \\  
        Range & 52.9$\approx$50.0 & 13.74$\rightarrow$12.00\dr{12.73\%} & 27.45$\rightarrow$27.35\dr{0.36\%} \\  
        Variate & 56.6$\approx$50.0 & 3.69$\rightarrow$3.06\dr{17.0\%} & 51.16$\rightarrow$54.16\ug{5.53\%} \\
        \bottomrule
    \end{tabular}
    }
    \caption{C1: ViTST performs close to random guessing. 
    C2: LLMs demonstrate moderate robustness. 
    C3: MLLMs remain robust under irregular conditions.
    "$\rightarrow$" denotes from regularity to irregularity.}
    \vspace{-0.8cm}
    \label{tab:hypotheses}
\end{table}

\begin{figure}[b!]
    \vspace{-0.3cm}
    \centering
    \includegraphics[width=0.48\textwidth]{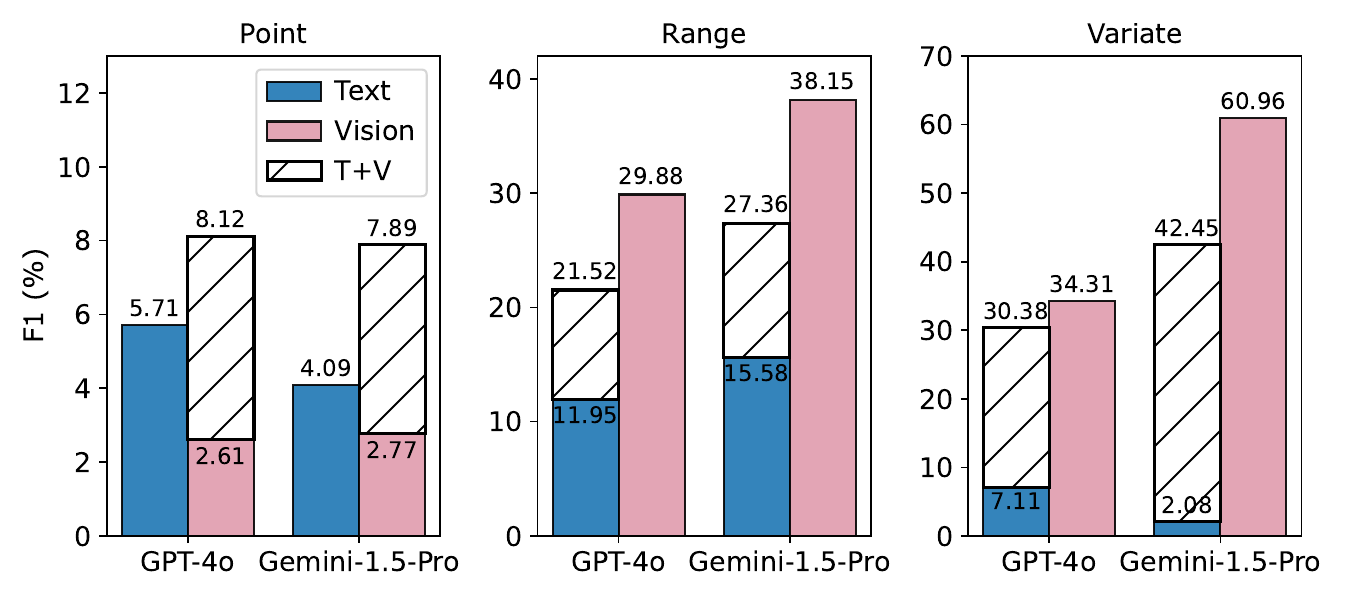}
     \caption{Impact of input modality on performance. 
     When the input time series changes from text to image, MLLMs show a performance drop for point-wise anomalies but improved effectiveness for range- and variate-wise anomalies. When both text and image are provided as input (\textit{T+V}), performance is generally between \textit{text} or \textit{image} alone.
     }
     \vspace{-0.7cm}
    \label{fig:bar_input_modality}
\end{figure}

\textbf{Origin of Resilience: Visualization or LLM?} 
MLLMs remain robust under irregular sampling, but the source of this resilience, visualization, LLM reasoning, or both, remains unclear. 
To examine this, we test three conditions: 
(C1) visualization alone (without LLMs), 
(C2) LLMs alone (without visualization), 
and (C3) a combination of both visualization and LLMs. 

To test C1, we use ViTST~\cite{li2023time}, a pre-trained vision Transformer~\cite{liu2021swin} effective for irregular time-series classification. 
A direct application to TSAD would require pixel-level anomaly labels ($\approx$4.4k images), which is prohibitively expensive. 
Instead, we simplify the task to binary classification: given a visualized irregular TSI, predict whether it contains anomalies. 
If ViTST cannot achieve reliable accuracy, it likely cannot localize anomalies. 
As shown in Table~\ref{tab:hypotheses}, ViTST performs close to random guessing ($\approx$50\% accuracy), indicating that \textit{visualization alone} is insufficient for detecting anomalies under irregular conditions. To test C2, we treat time series as textual data, replacing missing points with “NaN” tokens and explaining them in the prompt, thereby activating only their language modeling capability. 
Comparing regular and irregular inputs in text mode (\textit{MLLM (Text)}), Table~\ref{tab:hypotheses} shows that \textit{LLMs alone} exhibit moderate robustness. For example, the F1 score for range anomalies drops from 13.74 to 12.00, representing a 12.73\% decrease.
Finally, for C3, we evaluate MLLMs on visual inputs (\textit{MLLM (Vision)}). 
Their performance remains stable on irregular data, confirming that \textbf{the synergy between visualization and LLM reasoning provides the strongest resilience to irregularity}.

\section{RQ3: Input Modality in TSAD}
To investigate RQ3, we evaluate three input formats: \textit{text} (encoding time series as text), \textit{vision} (converting time series into images), and their combination (\textit{T+V}).
\\\textbf{Input Modality on Performance.}
Figure~\ref{fig:bar_input_modality} shows that shifting from textual to visual input leads to a performance drop for point- anomalies (e.g., GPT-4o: 5.71 $\rightarrow$ 2.61) but yields consistent improvements for range- and variate-wise anomalies (e.g., GPT-4o: 11.95 $\rightarrow$ 29.88). 
The results indicate \textbf{an information focus shifts between text and image: text emphasizes quantitative variations while images highlight qualitative patterns}. The textual representation makes numerical deviations explicit, which is advantageous for detecting subtle point-wise anomalies. In contrast, the visual representation amplifies coarse-grained temporal structures, benefiting range- and variate-wise anomaly detection.

When both modalities are provided, the T+V setting yields the best performance on point-wise anomalies, due to the complementary nature of textual and visual information. In contrast, for range- and variate-wise anomalies, the visual (V) modality alone outperforms others (T and T+V). This discrepancy may be attributed to long-context learning challenges~\cite{li2024long} of LLMs.
For instance, the length for a univariate series is 400, while a multivariate series is up to 20×200. This issue is particularly evident in the low F1 scores for variate-wise anomalies under the T setting.

\begin{table}[b!]
    \centering
    \scalebox{1.0}{
    \begin{tabular}{lcrcc}
        \toprule
        & Text & Vision & T+V \\
        \midrule
        Llava-Next-72B & 98.3$\pm$0.6 & 21.7$\pm$1.5\dg{76.6} & 59.7$\pm$1.1\dg{38.6} \\
        Llava-Next-8B & 100.0$\pm$0.0 & 76.3$\pm$1.2\dg{23.7} & 86.7$\pm$0.9\dg{13.3} \\
        Qwen2-VL-72B & 13.3$\pm$1.2 & 0.0$\pm$0.0\dg{13.3} & 9.7$\pm$2.6\dg{3.6} \\
        Qwen2-VL-7B & 98.7$\pm$0.3 & 0.0$\pm$0.0\dg{98.7} & 97.0$\pm$0.0\dg{1.7} \\
        GPT-4o-mini & 100.0$\pm$0.0 & 0.0$\pm$0.0\dg{100.0} & 56.6$\pm$0.3\dg{43.4} \\
        Gemini-1.5-Flash & 100.0$\pm$0.0 & 0.0$\pm$0.0\dg{100.0} & 100.0$\pm$0.0\dg{0.0} \\
        \bottomrule
    \end{tabular}
    }
    \caption{Impact of input modality on MLLMs' hallucinations. Each cell denotes the number of hallucinations out of 100 samples. Compared to the text modality (\textit{T}), the image modality (\textit{V}) significantly reduces hallucinations. When both modalities exist (\textit{T+V}), hallucination occurrence typically fall between \textit{text} or \textit{image} alone.
    }
    \vspace{-0.5cm}
    \label{tab:hallucination}
\end{table}

\textbf{Input Modality on Hallucinations.}
We find that MLLMs, particularly small-scale models, are prone to generating hallucinated responses, and input modality does influence them. For example, when given a 36-dimensional time series as input, LLaVA-NeXT-8B may produce an ungrounded sequence such as "[0, 1, 2, \dots, 100, \dots]" until reaching the maximum token limit. Table~\ref{tab:hallucination} shows the frequency of hallucinations occurrence across MLLMs. The results reveal a consistent observation: \textbf{visual input substantially reduces hallucination rates compared to text}. For example, Llava-Next-72B has 98.3 hallucinations with text input but only 21.7 with visual input. Still, we attribute it to \textbf{lengthy input textual series}. It can be confirmed by the fact that \textit{T+V} modality does not reduce hallucinations compared to \textit{Vision} in the Table~\ref{tab:hallucination}.

\begin{figure*}[t]
    \centering
    \includegraphics[width=0.95\textwidth]{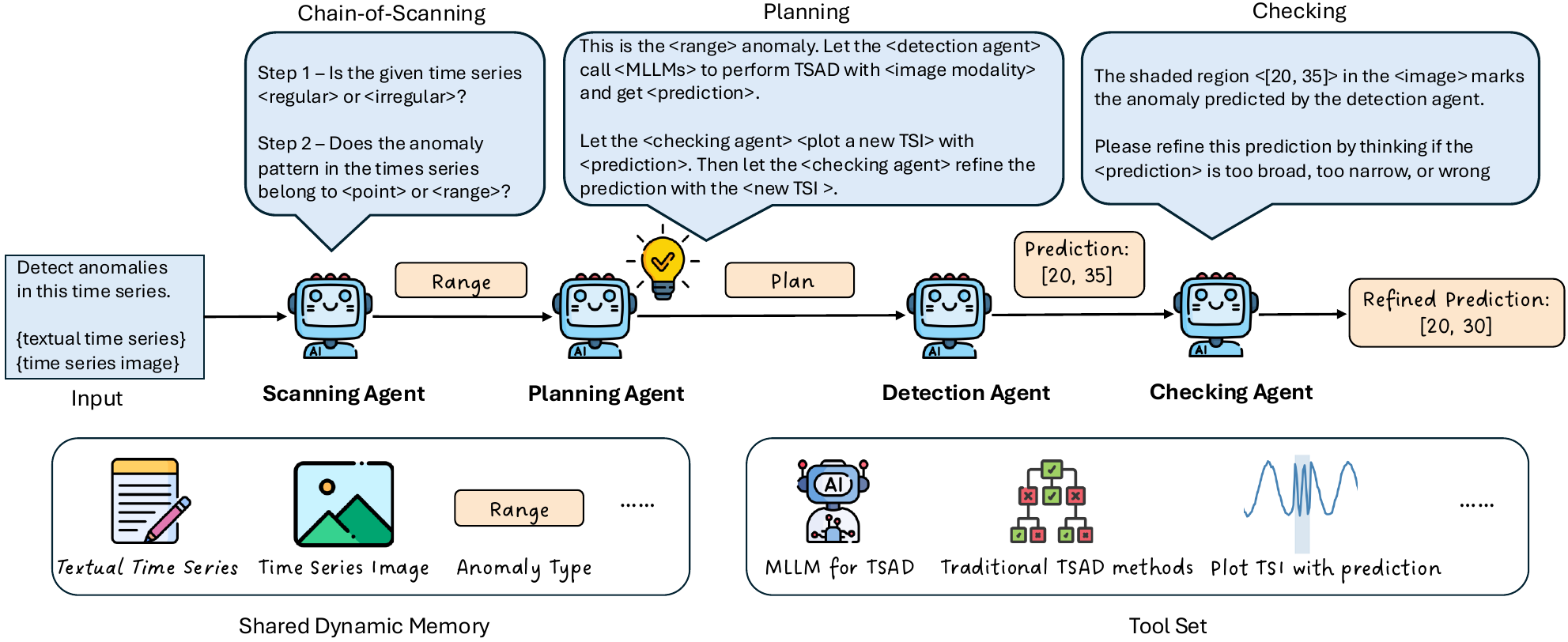}
    \caption{The overview of {\m}. The scanning agent reasons about the anomaly type. Based on the anomaly type, the planing agent develops a customized plan to guide the following detection and checking agents. The detection agent executes anomaly detection by calling appropriate methods from a tool set. The checking agent finally refines the prediction. A dynamic memory to store key messages is shared by all agents. The {\m} adaptively integrates the strengths of traditional methods (effective at point-wise anomalies) and MLLMs (powerful for range-wise and irregular anomalies), and also leverages the MLLMs' inherent capabilities like reasoning, planning, tool-use, and self-reflection.}
    \label{fig:tsad-agents}
\end{figure*}

\section{TSAD-Agents}
Our empirical findings reveal that MLLMs struggle with detecting fine-grained anomalies. To address this limitation, we propose a multi-agent framework, {\m}, which systematically leverages these insights along with the inherent abilities of MLLMs, such as reasoning, planning, self-reflection, and tool-use capabilities. We present {\m} through its motivation, architectural design, and experimental evaluation. {\m} specifically targets point- and range-wise anomalies, as our findings indicate that MLLMs are already effective at detecting variate-wise anomalies. 

\vspace{-0.3cm}
\subsection{Motivation}
Existing TSAD methods often make strong assumptions about the form of anomalies, i.e., assuming anomalies are either points or ranges. For example, iForest~\cite{liu2008isolation} is designed to detect point-wise anomalies, whereas TranAD~\cite{tuli2022tranad} assumes range-wise anomalies. These assumptions require prior knowledge of the anomaly type and may not hold in real-world scenarios where anomalies can manifest in diverse or unforeseen ways. Additionally, time series can be collected in unstable environments, resulting imperfect time series with missing values.

To address these issues, we propose {\m}, a novel framework built upon MLLMs to enable automatic TSAD without any prior assumption on anomaly types. {\m} is designed to adaptively handle different anomaly forms and missing data by leveraging the complementary strengths of both traditional TSAD algorithms and modern MLLM-based analysis. This framework aims to eliminate the need for manual selection of anomaly type assumptions, making anomaly detection more robust and applicable to real-world time series data.

\begin{table*}[t]
\centering
\scalebox{0.80}{
\begin{tabular}{l|cccccccccccc}
\hline
\textbf{} & \multicolumn{3}{c}{\textbf{Point}} & \multicolumn{3}{c}{\textbf{Range}} & \multicolumn{3}{c}{\textbf{Irr-Point}} & \multicolumn{3}{c}{\textbf{Irr-Range}} \\
\textbf{} & \textbf{P} & \textbf{R} & \textbf{F1} & \textbf{P} & \textbf{R} & \textbf{F1} & \textbf{P} & \textbf{R} & \textbf{F1} & \textbf{P} & \textbf{R} & \textbf{F1} \\ \hline
iForest & 44.4 ± 0.6 & 97.7 ± 1.1 & 62.1 ± 0.9 & 11.5 ± 0.0 & 64.6 ± 1.0 & 19.7 ± 0.4 & 16.0 ± 0.2 & \textbf{98.8 ± 1.0} & 27.2 ± 0.3 & 7.9 ± 0.0 & \textbf{76.7 ± 1.5} & 14.0 ± 0.0 \\
OCSVM & 10.0 ± 0.1 & 97.1 ± 0.3 & 18.0 ± 0.1 & 6.3 ± 0.1 & 62.5 ± 1.5 & 11.3 ± 0.1 & 7.5 ± 0.1 & 96.4 ± 1.5 & 14.1 ± 0.1 & 5.6 ± 0.0 & 69.3 ± 0.2 & 10.2 ± 0.1 \\ \hline
OmniAnomaly & 5.2 ± 0.1 & 83.3 ± 1.8 & 9.7 ± 0.1 & 8.5 ± 0.1 & \textbf{99.3 ± 0.3} & 15.6 ± 0.1 & 4.2 ± 0.0 & 34.8 ± 0.0 & 7.5 ± 0.1 & 7.1 ± 0.1 & 25.8 ± 0.2 & 11.2 ± 0.1 \\
THOC & 3.6 ± 0.1 & 73.0 ± 1.8 & 7.1 ± 0.1 & 14.8 ± 0.1 & 70.0 ± 1.7 & 24.6 ± 0.4 & 3.4 ± 0.0 & 29.2 ± 0.6 & 6.1 ± 0.0 & 12.4 ± 0.3 & 35.7 ± 0.4 & 17.9 ± 0.1 \\
TranAD & 5.3 ± 0.1 & \textbf{97.9 ± 1.0} & 10.1 ± 0.1 & 5.5 ± 0.1 & 94.6 ± 1.8 & 10.4 ± 0.1 & 4.9 ± 0.1 & 23.6 ± 0.2 & 8.2 ± 0.1 & 7.9 ± 0.1 & 13.3 ± 0.1 & 10.0 ± 0.0 \\ \hline
Prompting & 11.2 ± 0.5 & 2.2 ± 0.2 & 3.5 ± 0.2 & 23.4 ± 0.2 & 61.4 ± 1.5 & 32.9 ± 1.1 & 10.6 ± 0.1 & 1.4 ± 0.0 & 2.4 ± 0.1 & 27.2 ± 0.5 & 45.2 ± 1.1 & 30.8 ± 0.3 \\
LLMAD & 15.3 ± 0.8 & 3.2 ± 0.2 & 5.4 ± 0.5 & 28.1 ± 0.3 & 25.1 ± 0.2 & 26.7 ± 0.5 & 13.6 ± 0.2 & 3.0 ± 0.0 & 4.8 ± 0.0 & 27.8 ± 0.5 & 19.2 ± 0.3 & 22.6 ± 0.2 \\
SIGLLM & 14.0 ± 0.1 & 8.7 ± 0.1 & 10.7 ± 0.2 & 14.3 ± 0.1 & 16.1 ± 0.1 & 15.0 ± 0.1 & 5.1 ± 0.1 & 19.1 ± 0.2 & 8.2 ± 0.1 & 10.5 ± 0.2 & 18.4 ± 0.3 & 13.4 ± 0.2 \\ 
TAMA & 19.3 ± 0.3 & 7.1 ± 0.7 & 10.2 ± 0.4 & 19.9 ± 0.3 & 37.1 ± 0.2 & 25.9 ± 0.2 & 5.4 ± 0.4 & 10.7 ± 0.1 & 7.1 ± 0.2 & 8.9 ± 1.0 & 15.3 ± 1.6 & 11.4 ± 1.1  \\ \hline
TSAD-Agents & \textbf{55.4 ± 0.8} & 81.9 ± 0.5 & \textbf{65.4 ± 0.6} & \textbf{30.5 ± 0.2} & 53.5 ± 0.1 & \textbf{38.6 ± 0.5} & \textbf{17.6 ± 0.2} & 67.5 ± 1.0 & \textbf{29.0 ± 0.2} & \textbf{32.0 ± 0.4} & 41.0 ± 0.3 & \textbf{36.5 ± 0.5} \\
\hline
\end{tabular}
}
\caption{Performance comparison of {\m} and baselines. The results demonstrate the superiority of {\m}.}
\label{tab:tsad-agents}
\end{table*}

\subsection{Architecture}
Figure~\ref{fig:tsad-agents} shows that {\m} consists of four MLLM-driven modules that work together in an end-to-end loop: a Scanning agent, a Detection agent, and a Checking agent. These agents work synergistically to integrate the strengths of traditional TSAD methods and the multimodal reasoning capabilities of MLLMs.
\\\textbf{Scanning Agent} is responsible for initially reasoning the anomaly type in the input time series before performing detailed detection. We propose a \textit{Chain-of-Scanning (CoS)} technique that guides the agent to hierarchically infer the anomaly type step by step. 
\begin{itemize}
    \item Step 1: determine if the time series is regular or irregular;
    \item Step 2: determine if potential anomalies are point- or range-wise.
\end{itemize}
Based on this step-by-step reasoning analysis, the scanning agent classifies the anomaly into one of four categories: point, range, irr-point, or irr-range.
By autonomously determining the anomaly category, the Scanning Agent guides the subsequent detection approach without prior human assumption about the anomaly type.
\\\textbf{Planing Agent} generates a tailored plan to instruct the detection and checking agents, adapting to the inferred anomaly type. Drawing on empirical insights from the benchmark, the planning agent adopts different detection strategies and modalities:
\begin{itemize}
    \item Point: traditional methods (text modality) for detection; hybrid modality for checking.
    
    \item Range: MLLMs (image modality) for detection and checking.
    
    \item Irr-Point: traditional methods (text modality) for detection; hybrid modality for checking.
    
    \item Irr-Range: MLLMs (image modality) for detection and checking.
\end{itemize}
This planning stage comprehensively combines strengths of classical techniques (effective at point-wise anomalies) with MLLMs (strong at range-based patterns).
\\\textbf{Detection Agent} executes anomaly detection using a tool set that includes traditional TSAD methods (e.g., iForest) and MLLM-based detectors (e.g., Gemini) after receiving the guidance from the planning agent. This design ensures that {\m} can handle a wide spectrum of anomaly types automatically: it exploits the precision of classical methods for point outliers and the flexibility of advanced MLLMs for coarse-grained anomaly patterns.
\\\textbf{Checking Agent} is a verification and refinement module that embodies a form of self-reflection in the {\m} framework. After the detection agent predicts a set of anomalies, the checking agent evaluates these predictions to confirm their validity. It selects the appropriate tool to highlight the predictions, either via textual series or TSI plots. With the enriched representations, the checking agent is able to think if the original prediction is too broad or too narrow, and finally output refined prediction. This self-reflection mechanism can catch and correct its own errors, reduce false positives/negatives, and adaptively refine results without human intervention.
\\\textbf{Shared Dynamic Memory} is a central memory that dynamically stores key contextual information (e.g., anomaly type) inferred during the process. This memory can be retrieved by all agents, ensuring coherent and context-aware decision-making across modules. For example, after the scanning agent determines the anomaly type is range, this key message is stored in the memory, which will be accessed by the planning, detection and checking agents later.
\\\textbf{Tool Set} is a pool of tools accessible to the agents. The agents can select tools from this pool according to their needs. For example, the detection agent needs to use traditional methods in the tool pool to detect anomalies.

\subsection{Experimental Results and Ablation Study}
\textbf{Experimental Results.} We conduct comprehensive experiments to evaluate the effectiveness of {\m}. In our implementation, we include the traditional method iForest and Gemini-based detection methods in the tool set as they generally show better performance in the {\ds} findings. All agents are implemented using the LangChain framework and powered by Gemini. In addition to the baselines reported in the {\ds} benchmark, we incorporate recent LLM-based approaches for comparison: Prompting (Gemini detection baseline from {\ds}), LLMAD~\cite{liu2024large}, SIGLLM~\cite{alnegheimish2024large}, and TAMA~\cite{zhuang2024see}.

\begin{figure}[t]
    \centering
    \vspace{-0.5cm}
    \subfigure[F1]{
    	\begin{minipage}{0.23\textwidth}
   		 	\includegraphics[width=1\textwidth]{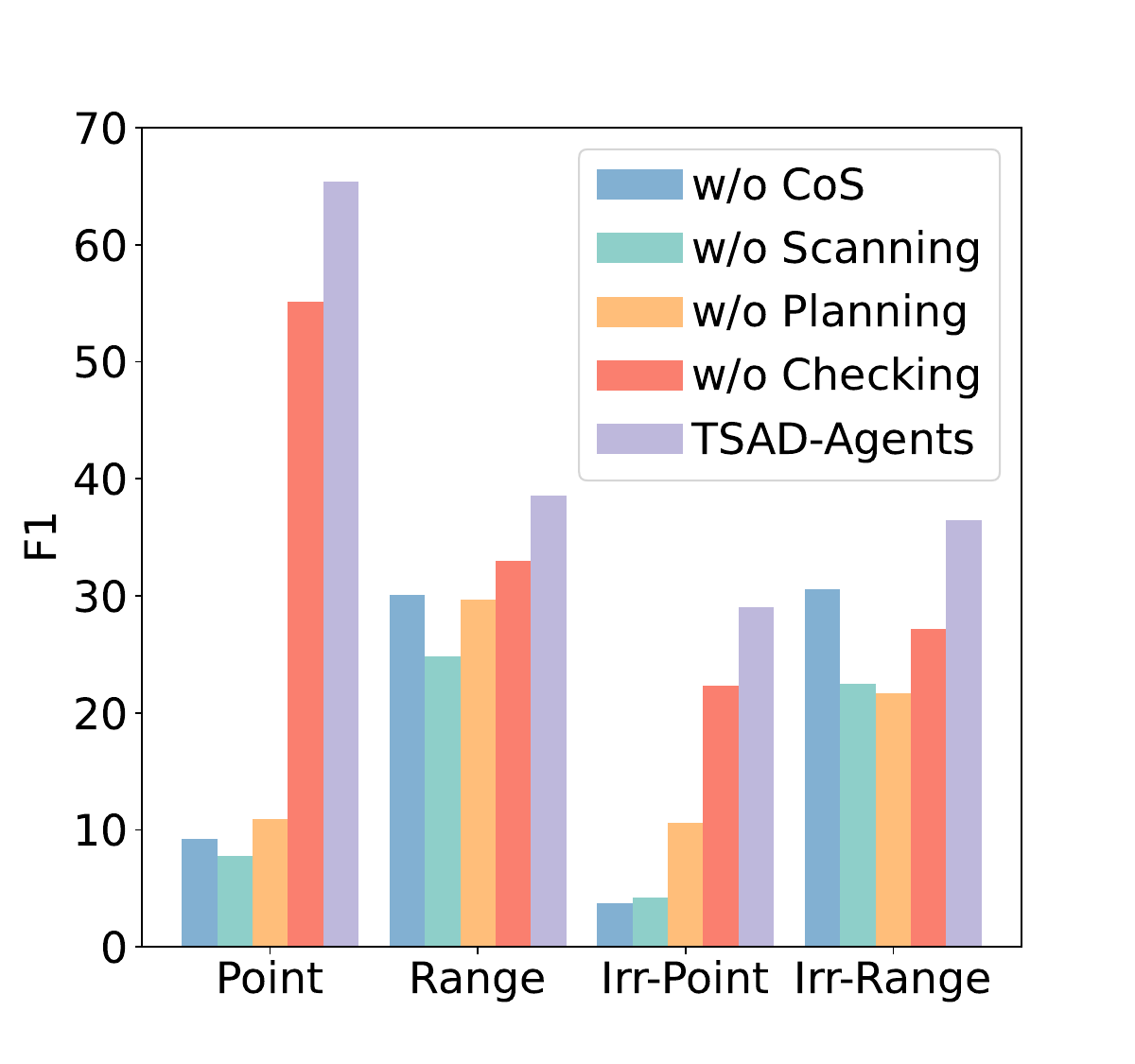}
    	\end{minipage}
    }
    \hspace{-0.3cm}
    \subfigure[Precision]{
    	\begin{minipage}{0.23\textwidth}
   		 	\includegraphics[width=1\textwidth]{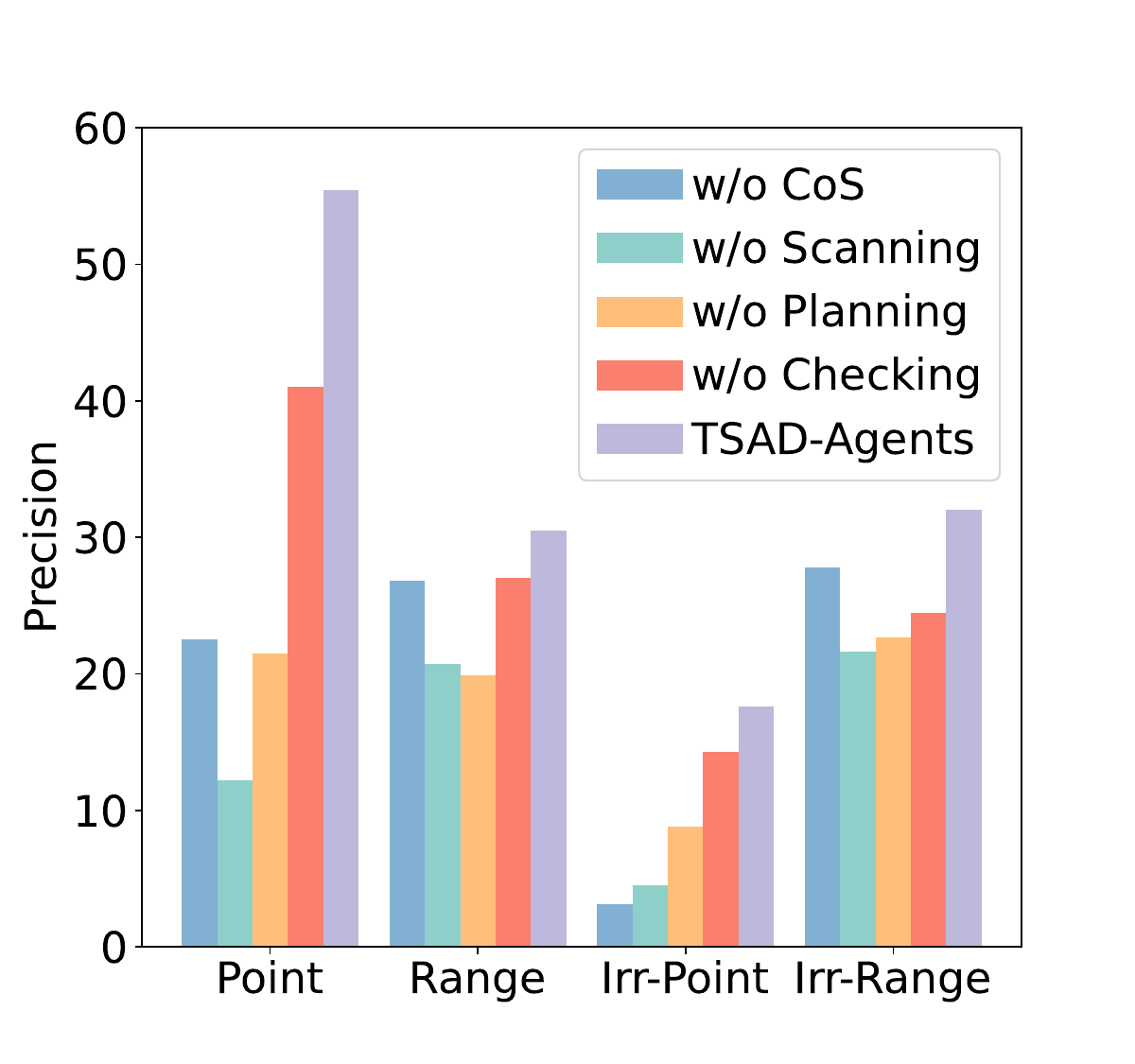}
    	\end{minipage}
    }
    \vspace{-0.4cm}
    \caption{Ablation study on for {\m}. The results indicate the effectiveness of each key component.}
    \label{fig:ablation}
    \vspace{-0.7cm}
\end{figure}

Table~\ref{tab:tsad-agents} demonstrates that {\m} consistently outperforms the baselines. Specifically, {\m} surpasses classical methods iForest on point anomalies, and outperforms the Prompting method on range anomalies. These results highlight {\m}'s ability to dynamically combine the precision of traditional TSAD methods (for point-wise anomalies) with the multimodal reasoning and self-reflection capabilities of MLLMs (for range-wise anomalies). We emphasize that this hybrid adaptive architecture allows {\m} to generalize effectively across diverse anomaly types, without prior specification on anomaly type, deliver state-of-the-art performance in both numeric and visual modalities.
\\\textbf{Ablation Study.} To verify the effectiveness of each key component of {\m}, including \textit{Chain-of-Scanning} technique, Scanning, Planning, and Checking agents, we design an ablation study on variants: w/o CoS, w/o Scanning, w/o Planning, and w/o Checking. Figure~\ref{fig:ablation} demonstrates that CoS technique and each agent is essential for {\m}. For example, eliminating the Planning Agent, which governs tool and modality selection, leads to a significant performance drop on both point- and range-wise anomalies, underscoring the importance of adaptive planning that bridges traditional methods and multimodal reasoning.

\section{Conclusion and Future Work}
In this paper, we investigate a research question: \textit{Can multimodal LLMs perform time series anomaly detection?} 
By building a early benchmark {\ds}, we offer several actionable insights of MLLMs for TSAD. Built on these insights, we propose the first multi-agent system for multimodal TSAD {\m}. With harnessing multiple powerful capabilities of MLLMs, TSAD-Agent enables automatic and accurate detection across diverse anomaly types and time series conditions. The future work can explore a more effective way to represent time series beyond raw images and reduce MLLMs' hallucinations.

\begin{acks}
This material is based upon work supported by NSF awards (SaTC-2241068, IIS-2506643, and POSE-2346158), a Cisco Research Award, and NSF NAIRR Pilot Award \#240469. The views and conclusions contained in this document are those of the authors and should not be interpreted as necessarily representing the official policies, either expressed or implied, of the National Science Foundation.
\end{acks}

\bibliographystyle{ACM-Reference-Format}
\bibliography{sample-base}

\appendix
\section{Related Work}
This work is related to two lines of research: (1) multimodal time series analysis, and (2) LLMs for time series anomaly detection.

\subsection{Multimodal Time Series Analysis} 
Time series analysis has evolved from traditional numerical processing~\cite{xu2024sst,lee2024z,du2024tsi} toward multimodal representations~\cite{xu2025beyond,liu2024time,pan2024s}. 
While time series were conventionally modeled as temporally ordered numerical sequences~\cite{rasul2023lag,das2023decoder,woo2024unified,goswami2024moment,ansari2024chronos}, recent advances demonstrate that they can be expressed through diverse modalities, such as text~\cite{zhou2023one,chang2023llm4ts,cao2023tempo,liu2024lstprompt,wang2024chattime}, images~\cite{zhang2023insight,daswani2024plots,cai2024timeseriesexam,zhong2025time,ma2025llm}, graphs~\cite{chen2023gatgpt,xu2023language,yan2024llm,qin2024llm,moghadas2024strada}, audio~\cite{yang2021voice2series}, and tables~\cite{wang2024tabletime,hoo2025tabular}.  
For instance, TimeGPT~\cite{garza2023timegpt} builds a time series foundation model pre-trained on large-scale sequential numerical data. PromptCast~\cite{xue2023promptcast} reformulates forecasting as a natural-language prompting task.  
ViTST~\cite{li2024time} finetunes a pre-trained vision transformer to perform medical classification by visualizing time series as images.  
In finance, multivariate time series can be represented as graphs~\cite{chen2023chatgpt} to capture inter-variate dependencies for stock movement prediction.  
Voice2Series~\cite{yang2021voice2series} reprograms pre-trained acoustic models for time series classification. 
TableTime~\cite{wang2024tabletime} reformulates multivariate time series classification as a table understanding task with Llama.
These works collectively highlight that the multimodality enables time series models to exploit cross-domain priors like language and vision.

\subsection{LLMs for Time Series Anomaly Detection}
The emergence of LLMs has inspired new paradigms for TSAD.  
LLMAD~\cite{liu2024large} leverages in-context learning and chain-of-thought reasoning to enhance few-shot anomaly detection.  
\cite{dong2024can} shows that prompting LLMs to think about the visual representation can assist anomaly identification.
Argos~\cite{gu2025argos} formalizes anomaly detection rules for interpretable anomaly detection via LLMs.  
Despite such progress, \cite{alnegheimish2024large} demonstrates that current LLM-based methods still under-perform traditional approaches by up to 30\% in F1 score, indicating that textual reasoning alone is insufficient for TSAD.  
Recent efforts thus explore multimodal LLMs (MLLMs) that integrate visual and temporal cues.  
As shown in the Table~\ref{tab:related}, \cite{zhou2025can} and \cite{zhuang2024see} are the most closely related studies: the former focuses on range-wise anomalies, treating point anomalies as special cases, whereas the latter emphasizes point-wise anomalies, viewing range anomalies as aggregated points.  
Different from prior work, our framework explicitly models \textit{point-wise}, \textit{range-wise}, and \textit{variate-wise} anomalies, and generalizes their regularity assumptions to handle \textit{irregular} time series.  
Building on these insights, we introduce a multi-agent MLLM framework that autonomously detects anomalies by activating the reasoning, planning, and self-reflective abilities of LLMs, without relying on prior assumption on anomaly granularity.

\section{The Details of Multimodal LLMs}\label{appendix:mllms}
We briefly describe the details of evaluated multimodal LLMs.
\\\textbf{GPT-4o}~\cite{achiam2023gpt} is a state-of-the-art multimodal model released by OpenAI. It is designed to handle both text and vision tasks with high efficiency, and optimized for complex reasoning tasks, achieving notable performance on benchmarks such as MGSM (math reasoning) and HumanEval (coding). Its architecture, while not publicly disclosed, is engineered to maintain consistent performance across modalities, ensuring that the inclusion of images does not degrade text-based reasoning capabilities.
\\\textbf{GPT-4o-mini}~\cite{achiam2023gpt} is a streamlined, cost-efficient variant of GPT-4o, designed to replace GPT-3.5 while surpassing its performance. It demonstrates strong capabilities in mathematical reasoning and coding. The model is not open-sourced and supports both text and vision inputs via its API.
\\\textbf{Gemini-1.5-Pro}~\cite{team2024gemini} is a high-performance model designed for advanced reasoning and multimodal tasks built by Google. It supports a large context window and excels in tasks requiring deep understanding, such as mathematical reasoning and code generation. Its architecture, while not publicly disclosed, ensures robust performance across text and vision inputs, maintaining consistency even when processing multimodal queries.
\\\textbf{Gemini-1.5-Flash}~\cite{team2024gemini} is a lightweight, high-speed variant of the Gemini-1.5 series, tailored for applications requiring rapid inference and low latency. Despite its compact design, it delivers competitive performance on benchmarks like MGSM and HumanEval. The model supports multimodal inputs, including text and vision, and can be called by API.
\\\textbf{LLaVA-NeXT-72B}~\cite{li2024llavanext-strong} is an advanced vision-language model tailored for high-performance multimodal understanding. It is open-source and built on a Qwen1.5-72B parameters architecture. It can integrate language and visual comprehension seamlessly, making it highly effective for image captioning, visual reasoning, and complex multimodal interactions. The model is designed to deliver state-of-the-art accuracy in vision-language tasks, with optimizations for both zero-shot and fine-tuned applications.
\\\textbf{LLaVA-NeXT-8B}~\cite{li2024llavanext-strong} is a smaller, more efficient version of the LLaVA-NeXT series, designed for applications where computational resources are limited. Built on LLaMA3-8B parameters, it maintains strong performance on vision-language tasks, achieving competitive results on benchmarks like MMMU and VQA. The model is particularly well-suited for real-time applications requiring fast inference and low memory usage.
\\\textbf{Qwen2-VL-72B-Instruct}~\cite{wang2024qwen2} is a large multimodal model optimized for instruction-following tasks involving both text and vision developed by Alibaba. With 72B parameters, it delivers robust performance on benchmarks such as MMMU and VQA. The model is designed to handle complex queries that require deep reasoning across modalities, ensuring consistent performance even when processing multimodal inputs. Its architecture is tailored for applications requiring high accuracy and reliability.
\\\textbf{Qwen2-VL-7B-Instruct}~\cite{wang2024qwen2} is a compact version of the Qwen2-VL series, designed for scenarios where computational efficiency is paramount. Despite its smaller size, it achieves competitive results on vision-language benchmarks, making it suitable for real-time applications. The model is optimized for instruction-following tasks, ensuring that it can handle multimodal inputs without compromising performance.

We evaluate the above MLLMs on eight NVIDIA RTX 1201
A6000 GPUs and set the temperature as 0.4 of MLLMs~\cite{zhou2025can}.

\section{Prompt Examples}\label{appendix:prompt}

\begin{figure}[h!]
\begin{AIbox}{Prompt for Point-Wise Time Series Anomalies}
{
    \textbf{Input Image}: 
    \begin{center}
        \includegraphics[width=0.8\textwidth]{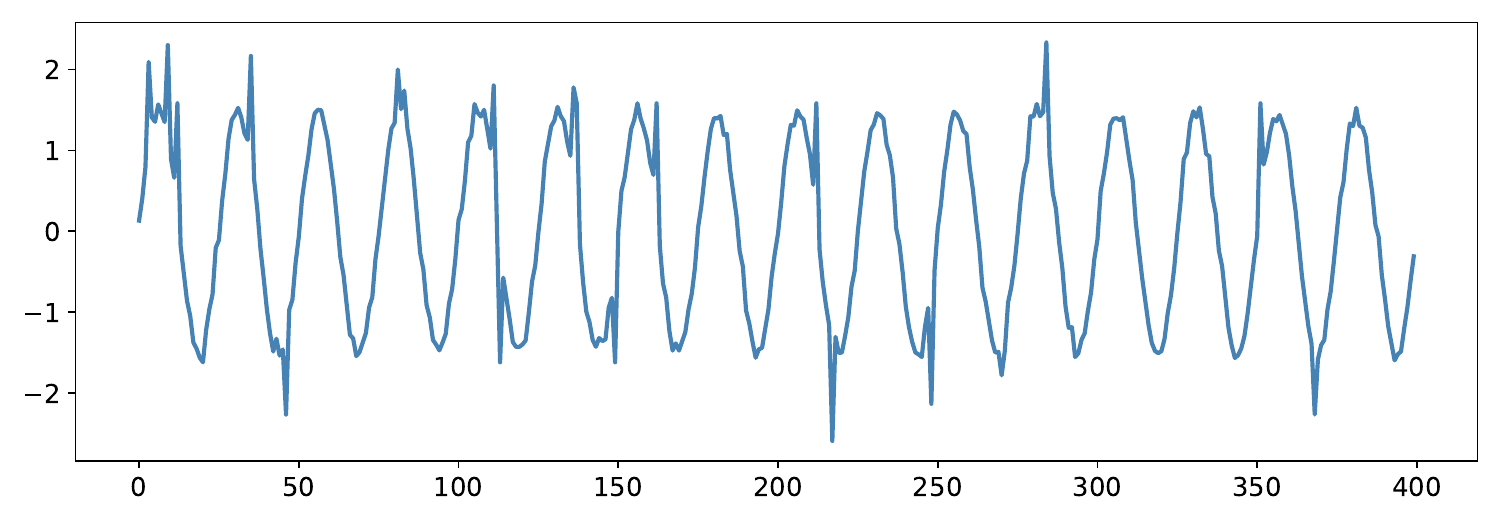}
    \end{center}
    \textbf{Prompt}: Detect points of anomalies in this time series, in terms of the x-axis coordinate. List one by one in a list. For example, if points x=2, 51, and 106 are anomalies, then output "[2, 51, 106]". If there are no anomalies, answer with an empty list [].
    
    \textbf{Response}: [3, 9, 12, 35, 46, 81, 83, 111, 113, 136, 137, 149, 162, 212, 217, 248, 270, 284, 351, 368]
}
\end{AIbox}
\end{figure}

\begin{figure}[h!]
\begin{AIbox}{Prompt for Range-Wise Time Series Anomalies}
{
    \textbf{Input Image}: 
    \begin{center}
        \includegraphics[width=0.8\textwidth]{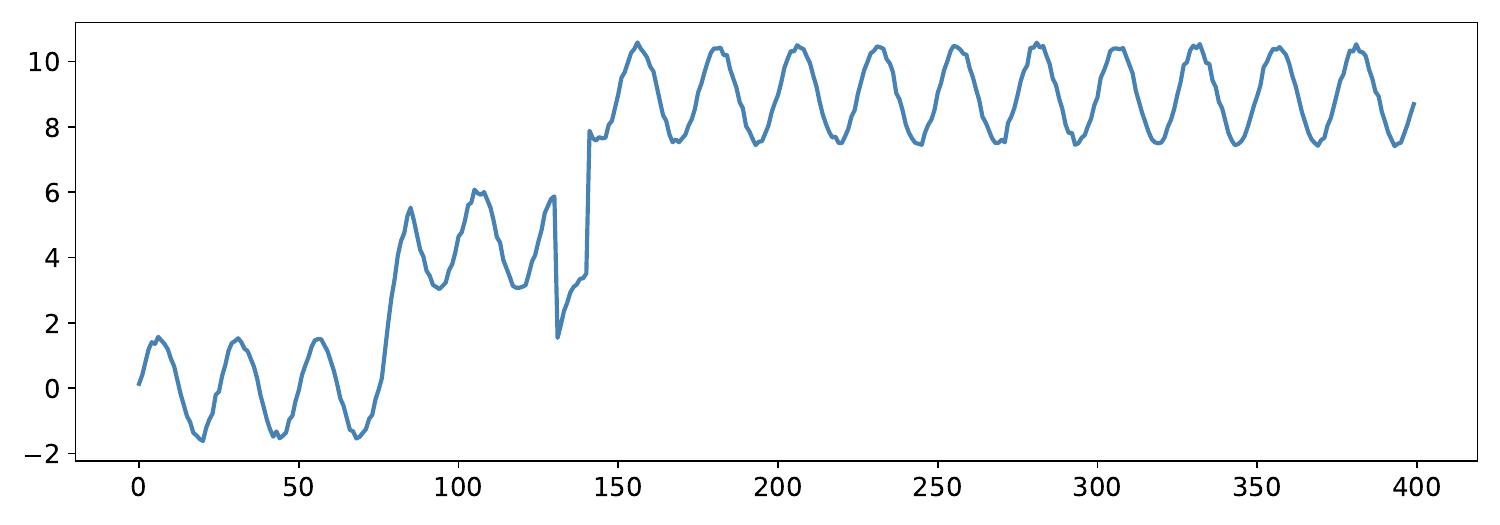}
    \end{center}
    \textbf{Prompt}: Detect ranges of anomalies in this time series, in terms of the x-axis coordinate. List one by one in a list. For example, if ranges (including two endpoints) [2, 11], [50, 60], and [105, 118], are anomalies, then output "[[2, 11], [50, 60], [105, 118]]". If there are no anomalies, answer with an empty list [].
    
    \textbf{Response}: [[76, 85], [131, 140]]
}
\end{AIbox}
\end{figure}

\begin{figure}[h!]
\begin{AIbox}{Prompt for Variate-Wise Time Series Anomalies}
{
    \textbf{Input Image}: 
    \begin{center}
        \includegraphics[width=0.8\textwidth]{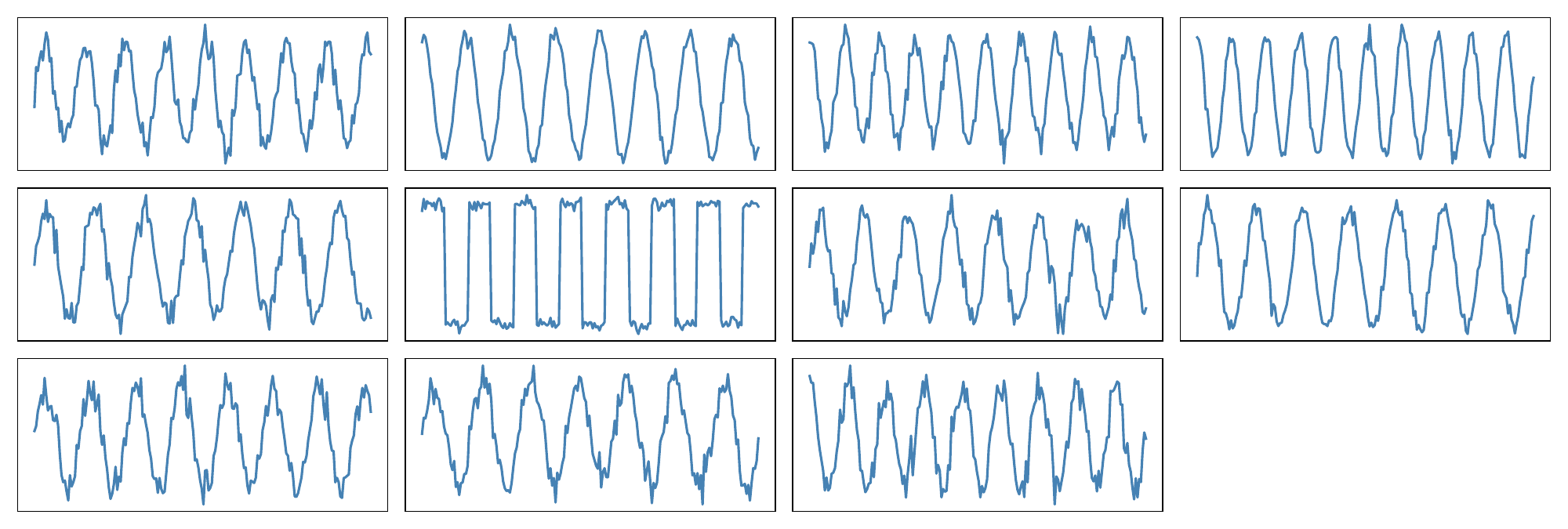}
    \end{center}
    \textbf{Prompt}: Detect univariate time series of anomalies in this multivariate time series, in terms of ID of univariate time series. The image is a multivariate time series including multiple subimages to indicate multiple univariate time series. From left to right and top to bottom, the ID of each subimage increases by 1, starting from 0. List one by one in a list. For example, if ID=0, 2, and 5 are anomalous univariate time series, then output "[0, 2, 5]". If there are no anomalies, answer with an empty list [].

    \textbf{Response}: [5]
}
\end{AIbox}
\end{figure}

\section{Solution for High-Dimensional Time Series}
A key challenge for MLLMs processing high-dimensional time series using image-based encoding is the reduced resolution of each variate, impairing models' performance. To address this, we propose a simple but effective splitting technique: we divide a high-dimensional time series image into multiple images, each covering a subset of variates. For example, a 100-dimensional time series can be represented as five 20-dimensional images. We conduct experiments with the splitting technique for GPT-4o and Gemini-1.5-Pro. The evaluation results across different time series dimensions are shown in Table~\ref{tab:dim}.

\begin{table}[t!]
\centering
\caption{The performance of MLLMs across different time series dimensions (anomaly ratio fixed at 5\%).}
\scalebox{0.80}{
\begin{tabular}{lccccc}
\toprule
\textbf{} & \textbf{20} & \textbf{40} & \textbf{60} & \textbf{80} & \textbf{100} \\
\midrule
\textbf{GPT} & 38.60 ± 0.45 & 36.42 ± 0.58 & 35.59 ± 0.33 & 34.21 ± 0.45 & 33.19 ± 0.55 \\
\textbf{Gemini} & 63.11 ± 0.68 & 62.16 ± 0.71 & 61.69 ± 0.48 & 60.87 ± 0.63 & 59.91 ± 0.66 \\
\bottomrule
\end{tabular}\label{tab:dim}
}
\end{table}

We observe that performance degrades slightly as dimensionality increases, which is expected due to the increased task complexity. However, the splitting approach significantly outperforms the vanilla method. For instance, Gemini-1.5-Pro achieves an F1 score of 59.91 at the dimension $M=100$ with the splitting technique, compared to only 33.2 at the dimension $M=36$ with the vanilla approach (Figure~\ref{fig:bar_multi}). This highlights the effectiveness of the splitting method in high-dimensional time series settings.

\section{The Impact of Number of Dimension $M$}\label{appendix:impact_M}

\begin{figure}[h!]
    \centering
    \includegraphics[width=0.48\textwidth]{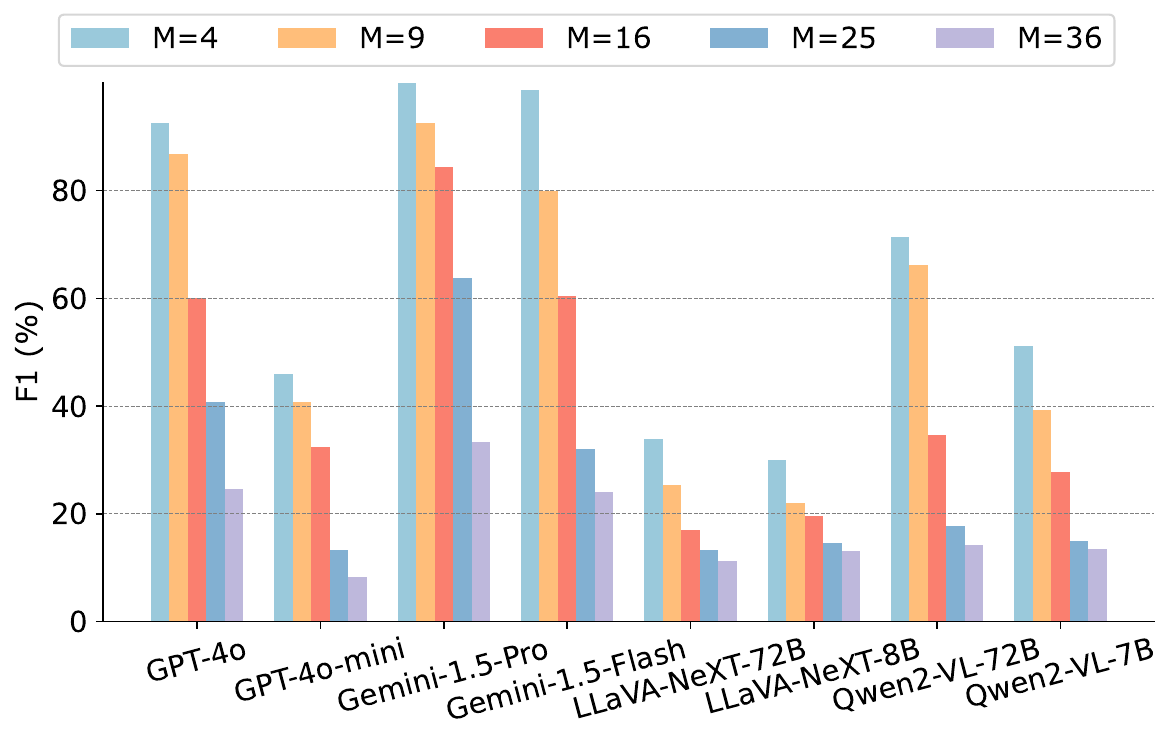}
     \vspace{-0.8cm}
     \caption{The impact of number of dimensions $M$ on MLLMs for variate-wise anomalies}
     \vspace{-0.5cm}
    \label{fig:bar_multi}
\end{figure}

We also investigate the impact of the number of variates $M$ in multivariate time series images. Intuitively, as $M$ increases, the information density that MLLMs must process grows. Moreover, this comes at the cost of reduced resolution for each subimage (variate). Consequently, \textbf{the effectiveness of MLLMs decreases as $M$ increases}. Figure~\ref{fig:bar_multi} illustrates this effect, showing a gradual decline in detection performance from $M=4$ to $M=36$, which aligns with our intuition. For instance, Gemini-1.5-Pro achieves an impressive F1 score of $100$ at $M=4$, but this drops significantly to $33.2$ when $M=36$. It demonstrates that MLLMs are struggling at high-dimensional multivariate time series.

\section{Implications on Time Series Anomaly Detection in the Era of Multimodal LLMs}\label{appendix:discussion}
Our empirical analysis and novel multi-agent framework offer important implications for TSAD and broader time series tasks in the era of multimodal LLMs.

First, our results suggest that \textbf{MLLMs and traditional TSAD methods can be flexibly and complementarily applied}. Their complementary strengths provide practical guidance on when to deploy MLLMs versus conventional approaches, depending on data granularities..

Second, the robustness of MLLMs to irregular time series, even with significant data dropped, highlights that \textbf{MLLMs have great potential in real-world applications with compromised data quality.} This resilience makes MLLMs a viable option for domains like healthcare, and IoT, where missing or irregularly sampled data is common. Organizations can rely on MLLMs to maintain detection accuracy despite these challenges, reducing the need for extensive data preprocessing or imputation.

Finally, our work points to \textbf{a paradigm shift in time series analysis: from purely numerical inputs toward multimodal representations}. By incorporating additional modalities (e.g., text, images, or domain knowledge), MLLMs can capture higher-level patterns that may be obscured in raw numerical formats. For instance, visual representations of time series can make qualitative patterns more pronounced, and effectively mitigate hallucinations compared to their textual counterparts. We encourage future research to further explore multimodal inputs across a broader range of classical time series tasks, including forecasting, classification, clustering, and emerging tasks.

\section{Hallucination Examples}\label{appendix:hallu}

\begin{figure}[h!]
\begin{AIbox}{Hallucination of LLaVA-NeXT-8B for 36-Variates Time Series Anomalies Detection (Random)}
{
    \textbf{Input Image}: 
    \begin{center}
        \includegraphics[width=0.75\textwidth]{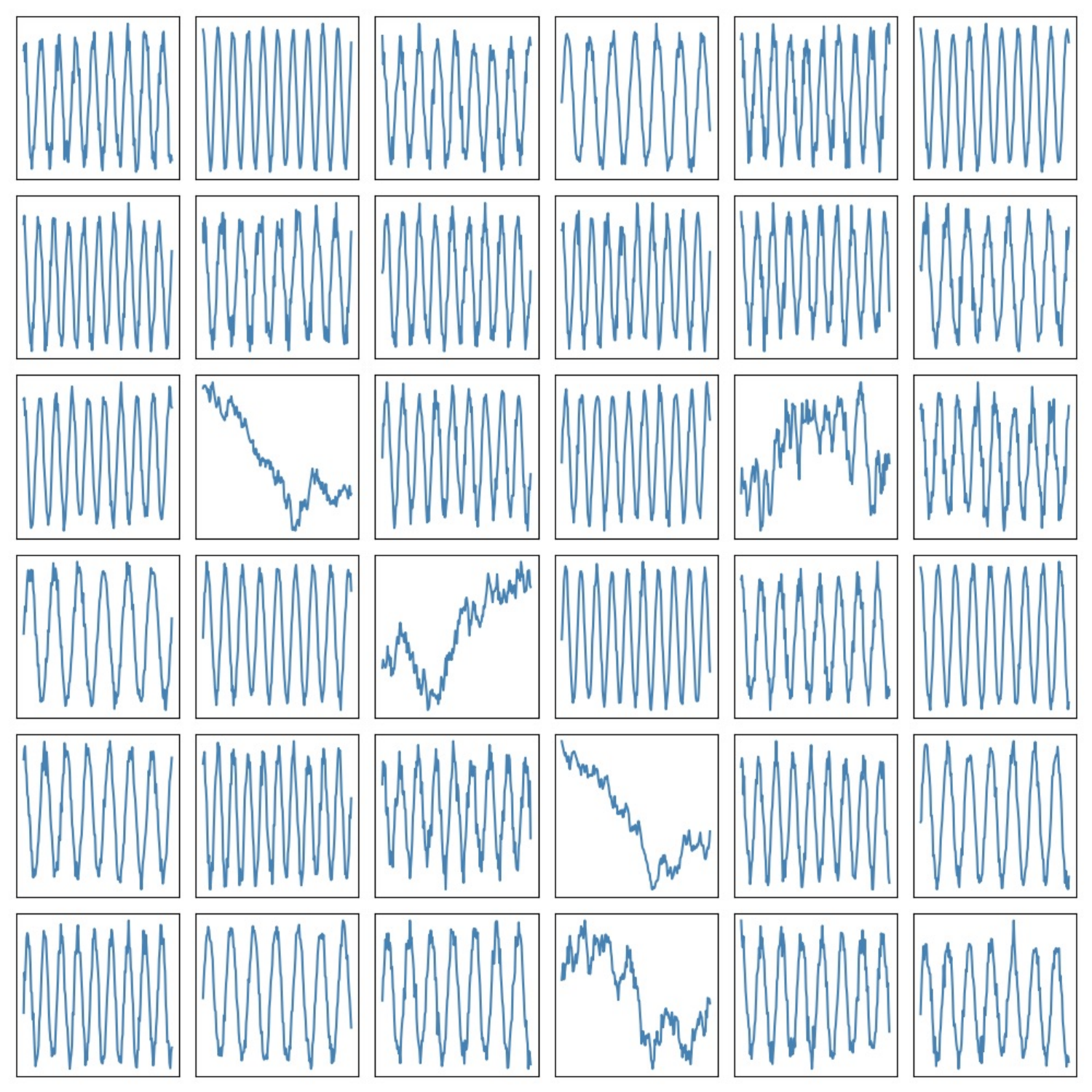}
    \end{center}
    \textbf{Prompt}: Detect univariate time series of anomalies in this multivariate time series, in terms of ID of univariate time series. The image is a multivariate time series including multiple subimages to indicate multiple univariate time series. From left to right and top to bottom, the ID of each subimage increases by 1, starting from 0. List one by one in a list. For example, if ID=0, 2, and 5 are anomalous univariate time series, then output "[0, 2, 5]". If there are no anomalies, answer with an empty list [].

    \textbf{LLaVA-NeXT-8B Response}: [0, 1, 2, 3, 4, 5, 6, 7, 8, 9, 10, 11, 12, 13, 14, 15, 16, 17, 18, 19, 20, 21, 22, 23, 24, 25, 26, 27, 28, 29, 30, 31, 32, 33, 34, 35, 36, 37, 38, 39, 40, 41, 42, 43, 44, 45, 46, 47, 48, 49, 50, \dots, 995, 996, 997]
}
\end{AIbox}
\end{figure}

\end{document}